%% file: main.tex
\documentclass[10pt,twocolumn,letterpaper]{article}

\usepackage{cvpr}      

\input{preamble}
\definecolor{cvprblue}{rgb}{0.21,0.49,0.74}
\usepackage[pagebackref,breaklinks,colorlinks,allcolors=cvprblue]{hyperref}

\usepackage{url}
\usepackage{graphicx}
\usepackage{multirow}
\usepackage{booktabs}
\usepackage{xcolor}
\usepackage{makecell}
\usepackage{caption}
\usepackage{cuted}
\usepackage{float}

\def\paperID{} 
\def\confName{CVPR}
\def\confYear{2027}

\title{\texttt{TextSphere}: Efficient Text-to-Image Generation via Shared Hyperspherical Alignment and Uniformity}

\author{Yixia Zhang \qquad
Girish A. Koushik \qquad
Xilu Wang \\
Department of Computer Science \\ University of Surrey\\
Guildford, United Kingdom\\
{\tt\small \{yixia.zhang,g.koushik,wangxilu\}@surrey.ac.uk}
\and
Wenwu Wang\\
CVSSP \\
University of Surrey\\
Guildford, United Kingdom\\
{\tt\small W.Wang@surrey.ac.uk}
}

\begin{document}
\maketitle

\input{sec/0_abstract}

\input{sec/1_intro}

\input{sec/2_related_work}
\input{sec/3_preliminaries}
\input{sec/4_method}

\input{sec/5_experiment}
\input{sec/6_conclusion}
{
    \small
    \bibliographystyle{ieeenat_fullname}
    \bibliography{main}
}

\input{sec/X_suppl}

\end{document}

%% file: preamble.tex
\usepackage{cuted} 

\usepackage{currfile} 

\usepackage{caption} 

%% file: sec/0_abstract.tex
\maketitle

\begin{strip}
    \centering
    \includegraphics[width=\textwidth]{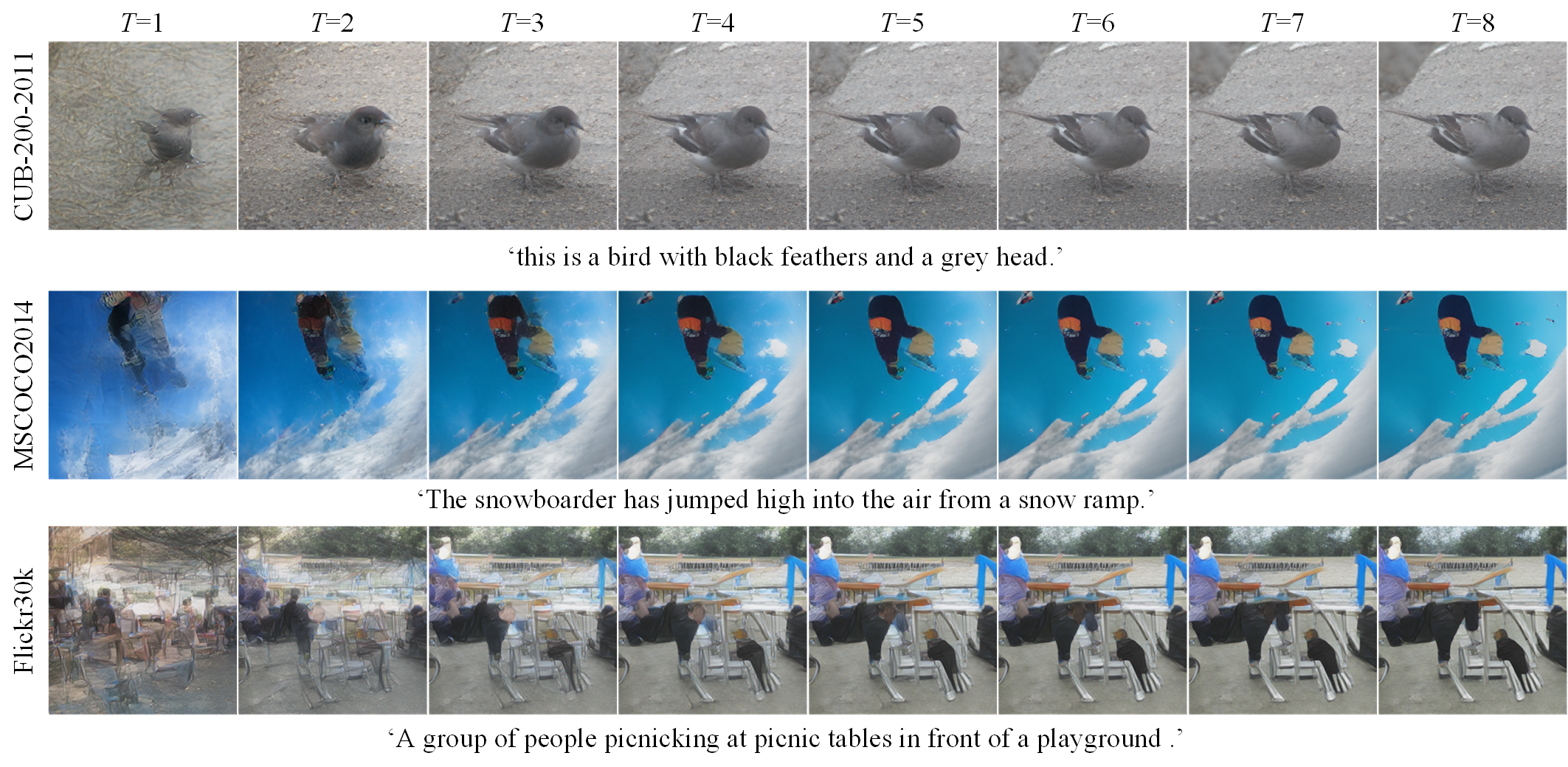}
    \captionsetup{hypcap=false}
    \captionof{figure}{Generated images using \texttt{TextSphere} over the first 8 refinement steps alongside corresponding textual descriptions across CUB-200-2011, MS-COCO-2014 and Flickr30k datasets.}
    \label{combine_sample_at_each_step}
\end{strip}


\begin{abstract}

Recent advances in text-to-image generation have substantially improved image quality, particularly through diffusion-based methods that require multiple inference steps. In contrast, Sphere Encoder offers an efficient generative framework capable of producing images with fewer forward steps. However, extending it to free-form text is nontrivial, because its class-conditioned design lacks cross-modal alignment. To address this challenge, we propose \texttt{TextSphere}, an efficient text-to-image generative framework that extends Sphere Encoder with language conditioning and cross-modal alignment. We incorporate pretrained CLIP text embeddings through adaptive layer normalization (AdaLN) to guide image generation while retaining the original spherical sampling mechanism. To bridge the cross-modal representation gap, we introduce a Spherical Semantic Alignment and Uniformity (SSAU) objective that jointly promotes semantic alignment and representation uniformity on the hypersphere. Experiments on CUB-200-2011, MS-COCO-2014, and Flickr30k demonstrate that \texttt{TextSphere} consistently reduces generation FID (gFID) over the text-conditioned baseline while achieving reconstruction FID (rFID) below 1.0. With only 4 and 6 sampling steps, \texttt{TextSphere} achieves lower gFID than the compared diffusion- and GAN-based models trained from scratch on CUB-200-2011 and MS-COCO-2014, respectively. We further observe that improved reconstruction fidelity does not necessarily yield better generation quality. In addition, we use Partial Least Squares Singular Value Decomposition (PLS-SVD) to analyze cross-modal alignment and modality gap, offering insight into the effectiveness of our approach.

\end{abstract}

%% file: sec/1_intro.tex

\section{Introduction}
\label{sec:intro}

Text-to-image (T2I) generation has made remarkable progress in recent years, enabling high-quality image synthesis from natural-language descriptions. Early approaches were primarily based on generative adversarial networks (GANs)~\cite{goodfellow2020generative}, while recent advances have shifted toward 
diffusion models~\cite{ho2020denoising}. These approaches, especially diffusion-based methods, have significantly improved image fidelity, semantic alignment, and high-resolution synthesis. However, generating high-quality images often requires multiple sampling steps, resulting in substantial computational overhead and inference latency. This motivates the development of efficient frameworks that can produce high-quality, text-aligned images in fewer steps.
For example, SANA-Sprint~\cite{chen2025sana} combines continuous-time consistency and latent adversarial distillation for ultra-fast generation, reducing inference steps from 20 to 1-4. Trajectory Distribution Matching~\cite{luo2025learning} and Score Identity Distillation~\cite{zhou2025few} enable flexible few-step generation through distribution matching. More recently, 1.x-Distill~\cite{li20261} explores fractional-step generation to balance image quality, diversity, and efficiency. Despite these advances, such methods typically rely on pretrained diffusion or flow-based teacher models and additional distillation procedures, introducing training complexity and computational overhead. This motivates the development of efficient generative frameworks that eliminate the need for pretrained generative teachers.

Recently, Sphere Encoder~\cite{yue2026image} introduced an efficient generative framework for class-conditioned image generation, achieving competitive image quality with fewer than 5 sampling steps. It learns to map images into a uniform spherical latent space through spherical normalization and reconstructs them using a decoder, enabling image generation by directly decoding randomly sampled spherical latents. Inspired by this work, we study the potential of using Sphere Encoder for text-conditioned image generation by directly replacing its original class embeddings with contrastive language-image pretraining (CLIP) text embeddings extracted from the corresponding captions, while retaining the original adaptive layer normalization (AdaLN) conditioning mechanism~\cite{perez2018film,peebles2023scalable}. To assess this adaptation, we employ Partial Least Squares Singular Value Decomposition (PLS-SVD)~\cite{zhu2026comet} to analyze the relationship between text and image representations. Our analysis in Section~\ref{pls_svd_analysis} reveals weak cross-modal covariance and directional correspondence, highlighting the limitations of direct text conditioning without explicit cross-modal alignment. We further investigate alternative conditioning mechanisms by comparing the cross-attention (CA) alignment approach~\cite{vaswani2017attention,rombach2022high} with the original AdaLN. Our preliminary experiments 
show that AdaLN consistently outperforms CA in generation quality within the Sphere Encoder framework
, consistent with recent findings in~\cite{hacohen2026ltx}.



Motivated by these observations, we propose \texttt{TextSphere}, an efficient text-to-image generative framework that extends Sphere Encoder with language conditioning and cross-modal alignment. Specifically, we incorporate pretrained CLIP text embeddings through AdaLN to guide image generation while retaining the original spherical sampling mechanism. To bridge the cross-modal representation gap, we introduce a Spherical Semantic Alignment and Uniformity (SSAU) objective that combines symmetric contrastive learning~\cite{radford2021learning} and spherical uniformity regularization~\cite{wang2020understanding}, jointly promoting semantic alignment and representation uniformity on the hypersphere.
The contrastive component encourages matched text–image representations to be closer while separating unmatched pairs, whereas uniformity regularization promotes well-distributed representations within each modality. By aligning textual representations with spherical image latents, our framework enhances cross-modal semantic correspondence while preserving the structured spherical latent space for image generation. To the best of our knowledge, this is the first work to extend the Sphere Encoder framework to text-to-image synthesis. Our main contributions are summarized as follows: 

\begin{itemize}
\item We propose \texttt{TextSphere}, an efficient text-to-image generative framework that extends Sphere Encoder with language conditioning and cross-modal alignment, enabling high-quality image synthesis with few sampling steps without relying on pretrained generative teacher models or distillation.


\item We propose an SSAU objective that combines symmetric contrastive learning~\cite{radford2021learning} and spherical uniformity regularization~\cite{wang2020understanding} to jointly promote semantic alignment and representation uniformity on the hypersphere, addressing the limitations of directly replacing class embeddings with text embeddings.




\item Extensive experiments on CUB-200-2011, MS-COCO-2014, and Flickr30k demonstrate consistent improvements in generation quality over the text-conditioned baseline. \texttt{TextSphere} also achieves competitive or superior gFID compared with diffusion- and GAN-based methods trained from scratch using only a few sampling steps. We further observe that improved reconstruction fidelity does not necessarily translate into better generation quality.

\end{itemize}

%% file: sec/2_related_work.tex
\section{Related Work}

Text-to-image generation has evolved from GANs to autoregressive and diffusion-based models. Early GAN-based approaches, such as AttnGAN~\cite{xu2018attngan}, DF-GAN~\cite{tao2022df}, and LAFITE~\cite{zhou2022towards}, explored text-guided image synthesis with efficient inference. Autoregressive models, including DALL-E~\cite{ramesh2021zero} and Parti~\cite{yu2022scaling}, demonstrated strong generation capabilities through sequential token prediction. More recently, diffusion-based methods, such as LDM~\cite{rombach2022high} and VQ-Diffusion~\cite{gu2022vector}, have achieved high-quality generation but typically require multiple sampling steps. To improve efficiency, approaches such as LCM~\cite{luo2023latent}, InstaFlow~\cite{liu2023instaflow}, and DMD~\cite{chen2026flash} enable few-step or single-step generation, often relying on pretrained models or distillation.

Unlike these approaches, Sphere Encoder~\cite{yue2026image} enables efficient image generation by directly sampling from a spherical latent space, without diffusion-based denoising or teacher distillation. Building upon this framework, we introduce an efficient text-to-image generative framework that extends Sphere Encoder with language conditioning and cross-modal alignment, enabling high-quality image synthesis with few sampling steps. A more comprehensive review of related work is provided in the Supplement. 

%% file: sec/3_preliminaries.tex
\section{Preliminaries}
\label{Preliminaries}

\paragraph{Sphere Encoder.} Sphere Encoder~\cite{yue2026image} is an encoder--decoder-based generative framework that maps images into a uniform spherical latent space through spherical normalization and reconstructs them using a decoder, enabling image generation by directly decoding randomly sampled spherical latents. Given an input image $\mathbf{x}\in\mathbb{R}^{H\times W\times 3}$, an encoder $E$ extracts a latent representation $\mathbf{z}=E(\mathbf{x})\in\mathbb{R}^{h\times w\times d}$. The latent representation is then projected onto a hypersphere through root mean square (RMS) normalization $f$:
\begin{equation}
\label{spherify}
    \mathbf{v}=f(\mathbf{z})
    =\sqrt{L}\frac{\mathbf{z}}{\|\mathbf{z}\|_2},
    \qquad L=hwd,
\end{equation}
where $L$ denotes the flattened latent dimension and $\|\mathbf{v}\|_2=\sqrt{L}$. A decoder $D$ reconstructs the image as $\hat{\mathbf{x}}=D(\mathbf{v})$.

During training, Sphere Encoder introduces Gaussian perturbations to the spherical latents to improve uniform latent-space coverage. Specifically, two perturbation levels are constructed:
\begin{equation}
\label{spherical_noise}
\begin{aligned}
    \mathbf{v}_{\mathrm{large}}
    &=f(\mathbf{v}+\sigma\boldsymbol{\epsilon}),\\
    \mathbf{v}_{\mathrm{small}}
    &=f(\mathbf{v}+\sigma_{\mathrm{sub}}\boldsymbol{\epsilon}),
\end{aligned}
\end{equation}
where the scalar $\boldsymbol{\epsilon}$ is uniformly sampled between [0, 1], $\sigma=r\sigma_{\max}$, $\sigma_{\max}$ is the maximum noise level. $\sigma_{\mathrm{sub}}=s\sigma$, and $s$ is uniformly sampled from [0, 0.5]. The model is jointly optimized using pixel reconstruction, pixel consistency, and latent consistency objectives:
\begin{equation}
\label{L_sphere}
\begin{aligned}
    \mathcal{L}_{\mathrm{Sphere}}
    =&\ \lambda_{\mathrm{recon}}\mathcal{L}_{\mathrm{recon}}\\
    &+\lambda_{\mathrm{pix}}\mathcal{L}_{\mathrm{pix\text{-}con}}\\
    &+\lambda_{\mathrm{lat}}\mathcal{L}_{\mathrm{lat\text{-}con}}, 
\end{aligned}
\end{equation}
\begin{equation}
\label{L_recon}
\begin{aligned}
    \mathcal{L}_{\mathrm{recon}} =\mathcal{L}_{L1+perceptual}(D(\mathbf{v}_{\mathrm{small}}),x),
\end{aligned}
\end{equation}
\begin{equation}
\label{L_pix-con}
\begin{aligned} 
    \mathcal{L}_{\mathrm{pix-con}} =\mathcal{L}_{L1+perceptual}(D(\mathbf{v}_{\mathrm{large}}),sg(D(\mathbf{v}_{\mathrm{small}}))), 
\end{aligned}
\end{equation}
\begin{equation}
\label{L_lat-con}
\begin{aligned}
    \mathcal{L}_{\mathrm{lat-con}} =\mathcal{L}_{cosine similarity}(v,E(D(\mathbf{v}_{\mathrm{large}}))), 
\end{aligned}
\end{equation}
where $\mathcal{L}_{L1+perceptual}$ combines smooth L1 loss~\cite{ren2016faster} and perceptual loss~\cite{johnson2016perceptual}, and $sg()$ denotes stop-gradient operation.


\paragraph{PLS-SVD.} Partial Least Squares Singular Value Decomposition is a statistical method for analyzing shared structures between modalities. Introduced in COMET~\cite{zhu2026comet}, PLS-SVD identifies projection directions that maximize cross-modal covariance, revealing shared semantic subspaces and representation gaps.

Given $N$ paired text and image embeddings, let $\mathbf{T},\mathbf{I}\in\mathbb{R}^{N\times C}$ denote their mean-centered representation matrices, where $C$ is the embedding dimension. PLS-SVD seeks projection directions $\mathbf{u}$ and $\mathbf{v}$ that maximize cross-modal covariance:
\begin{equation}
\max_{\|\mathbf{u}\|_2=\|\mathbf{v}\|_2=1}
\mathbf{u}^{\top}\mathbf{T}^{\top}\mathbf{A}\mathbf{v}.
\end{equation}
This is solved by applying SVD to the cross-modal covariance matrix $\mathbf{M}=\mathbf{T}^{\top}\mathbf{A}$:
\begin{equation}
\mathbf{M}=\mathbf{U}\boldsymbol{\Sigma}\mathbf{V}^{\top},
\end{equation}
where $\mathbf{U}$ and $\mathbf{V}$ contain the projection directions for each modality, and $\boldsymbol{\Sigma}$ contains singular values measuring their shared covariance. Larger singular values indicate stronger cross-modal dependencies, while directional correspondence between $\mathbf{U}$ and $\mathbf{V}$ provides additional insight into representation alignment. In our framework, we use PLS-SVD for post-training analysis, examining singular values and their corresponding projection directions to assess how well the proposed SSAU objective enhances text–image alignment.

%% file: sec/4_method.tex
\begin{figure*}[!t]
\centering
\includegraphics[width=17.4cm]{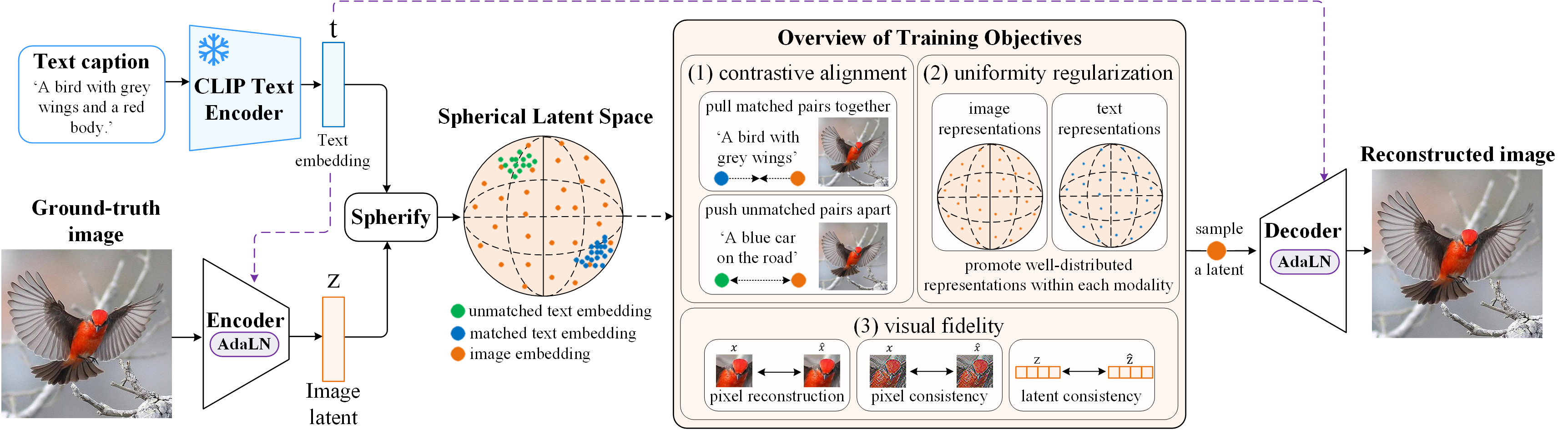}
\caption{\texttt{TextSphere} Framework Overview. It consists of a frozen pretrained CLIP text encoder, an image encoder, and an image decoder. During training, the image encoder maps ground-truth images onto a spherical latent space via RMS normalization (Spherify), as defined in Eq.~(\ref{spherify}), while the CLIP text encoder extracts semantic embeddings $\mathbf{t}$ from the corresponding captions. The training objectives jointly optimize (1) contrastive alignment, (2) uniformity regularization, and (3) visual fidelity, with the decoder sampling random latent representations from the hypersphere to generate images.}
\label{framework}
\end{figure*}

\section{Method}
\label{Method}




Fig.~\ref {framework} illustrates an overview of our framework. 
Based on our preliminary findings 
that AdaLN outperforms cross-attention (CA) (Table~\ref{ca_adaln_compare_cub} in the Supplement), we retain the original AdaLN conditioning mechanism to incorporate text embeddings into both the image encoder and decoder. The original Sphere Encoder employs reconstruction and consistency objectives, defined in Eq.~(\ref{L_sphere}), to preserve visual fidelity through pixel reconstruction, pixel consistency, and latent consistency. Motivated by the weak cross-modal correspondence revealed by our PLS-SVD analysis in Section~\ref{pls_svd_analysis}, we introduce the SSAU objective, which combines symmetric contrastive learning with spherical uniformity regularization to jointly enhance text-image semantic alignment and representation uniformity on the hypersphere.


\subsection{Spherical Semantic Alignment and Uniformity (SSAU)}



Given a batch of $B$ image–text pairs, let $I \in \mathbb{R}^{B \times D}$ and $T \in \mathbb{R}^{B \times D}$ denote the image representations and text embeddings extracted from the frozen CLIP text encoder, respectively, where $D$ is the feature dimension. For text-image alignment, both representations are $\ell_2$-normalized onto the unit hypersphere:
\begin{equation}
    \mathbf{u}_i = \frac{I_i}{\|I_i\|_2}, \qquad
    \mathbf{v}_i = \frac{T_i}{\|T_i\|_2},
\end{equation}
where $\mathbf{u}_i$ and $\mathbf{v}_i$ denote the normalized image and text representations of the pair $i$.

Following CLIP~\cite{radford2021learning}, we adopt a symmetric contrastive objective to encourage semantic alignment between image and text representations. Given the $\ell_2$-normalized image and text features, the temperature-scaled cosine similarity between the $i$-th image and $j$-th text is defined as
\begin{equation}
    S_{ij} = \frac{\mathbf{u}_i^\top \mathbf{v}_j}{\tau},
\end{equation}
where $\tau$ is a temperature parameter controlling the sharpness of the similarity distribution. We then compute the image-to-text and text-to-image contrastive losses using cross-entropy:
\begin{equation}
    \mathcal{L}_{\mathrm{i2t}} =
    -\frac{1}{B}\sum_{i=1}^{B}
    \log\frac{\exp(S_{ii})}
    {\sum_{j=1}^{B}\exp(S_{ij})},
\end{equation}
\begin{equation}
    \mathcal{L}_{\mathrm{t2i}} =
    -\frac{1}{B}\sum_{i=1}^{B}
    \log\frac{\exp(S_{ii})}
    {\sum_{j=1}^{B}\exp(S_{ji})}.
\end{equation}
The symmetric contrastive loss is defined as
\begin{equation}
    \mathcal{L}_{\mathrm{clip}} =
    \frac{1}{2}\left(
    \mathcal{L}_{\mathrm{i2t}}+
    \mathcal{L}_{\mathrm{t2i}}
    \right).
\end{equation}
This objective encourages matched image–text pairs to have higher similarity while separating unmatched pairs within the batch.

To complement the contrastive objective, we introduce spherical uniformity regularization~\cite{wang2020understanding} to encourage well-distributed representations within each modality. Given a batch of $\ell_2$-normalized representations $\mathcal{Z}=\{\mathbf{z}_i\}_{i=1}^{B}$, where $\mathbf{z}_i\in\mathbb{S}^{D-1}$, the uniformity loss is defined as
\begin{equation}
    \mathcal{L}_{\mathrm{uni}}(\mathcal{Z};t)
    =
    \log\left[
    \frac{1}{B(B-1)}
    \sum_{\substack{i,j=1\\i\neq j}}^{B}
    \exp\left(-t\|\mathbf{z}_i-\mathbf{z}_j\|_2^2\right)
    \right],
\end{equation}
where $t>0$ controls the sensitivity to pairwise distances. We apply this regularization separately to the normalized image and text representations:
\begin{equation}
    \mathcal{L}_{\mathrm{uniform}}
    =
    \frac{1}{2}
    \left[
    \mathcal{L}_{\mathrm{uni}}(\{\mathbf{u}_i\};t)
    +
    \mathcal{L}_{\mathrm{uni}}(\{\mathbf{v}_i\};t)
    \right].
\end{equation}
Note that although the pretrained CLIP text encoder is frozen, its outputs are further processed by trainable projection and feed-forward layers, allowing uniformity regularization to influence the representations of both modalities. This discourages embedding concentration, preserves representation diversity, and enhances semantic discriminability. 
The SSAU objective is defined as
\begin{equation}
\label{L_align}
    \mathcal{L}_{\mathrm{SSAU}}
    =
    \mathcal{L}_{\mathrm{clip}}
    +
    \lambda_u\mathcal{L}_{\mathrm{uniform}},
\end{equation}
where $\lambda_u$ controls the contribution of uniformity regularization. The overall training objective combines the original Sphere Encoder loss with the proposed SSAU objective:
\begin{equation}
\label{L_total}
    \mathcal{L}_{\mathrm{total}}
    =
    \mathcal{L}_{\mathrm{Sphere}}
    +
    \lambda_{\mathrm{SSAU}}\mathcal{L}_{\mathrm{SSAU}},
\end{equation}
where $\mathcal{L}_{\mathrm{Sphere}}$ is defined in Eq.~(\ref{L_sphere}) and $\lambda_{\mathrm{SSAU}}$ controls the alignment strength. By jointly optimizing 
pixel reconstruction, pixel and latent consistency, semantic alignment, and spherical uniformity, our framework promotes semantically aligned text–image representations while maintaining high visual fidelity in generated images.




\begin{figure*}[ht]
\centering
\includegraphics[width=13cm]{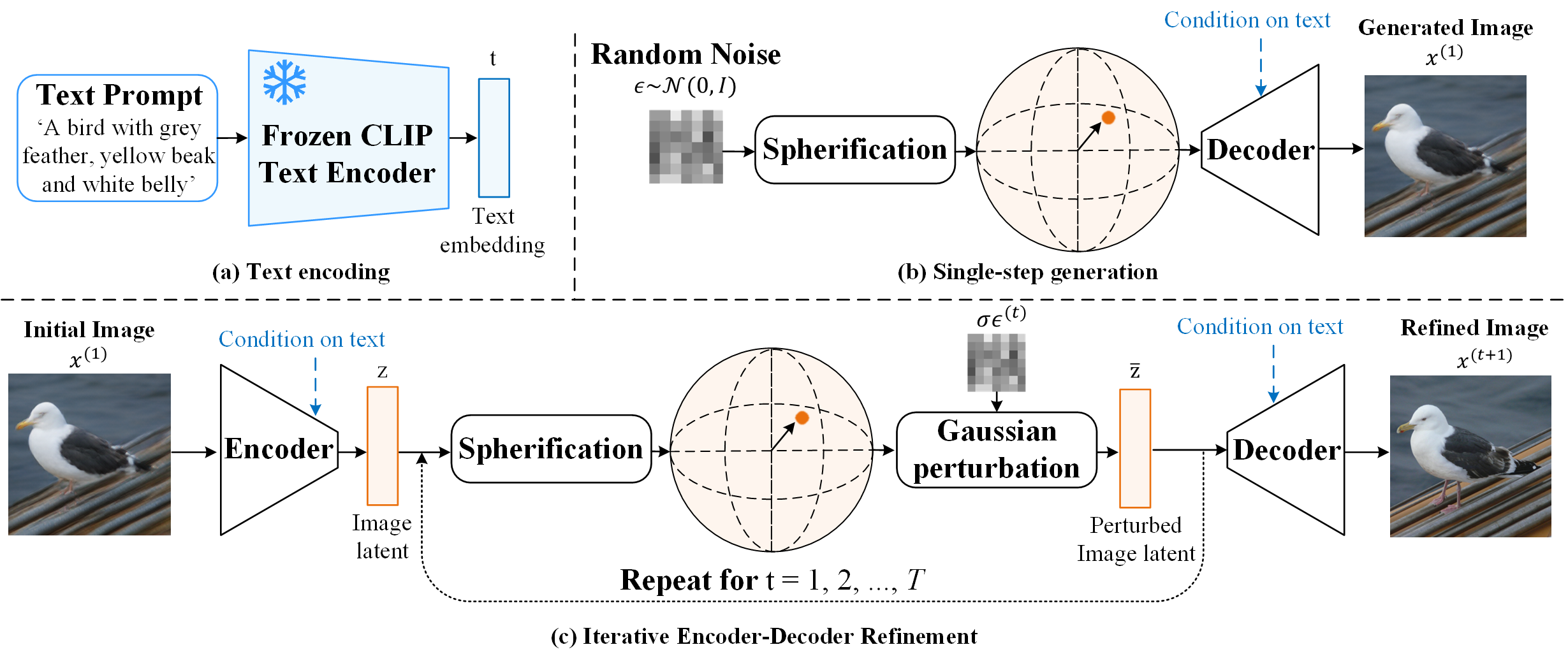}
\caption{Inference pipeline of TextSphere.}
\label{inference}
\end{figure*}

\subsection{Inference}
During inference, we perform text-conditioned sampling using a frozen pre-trained CLIP text encoder. The pipeline is presented in Fig.~\ref{inference}. Given a text prompt, its embedding is used to condition the generation process, as shown in Fig.~\ref{inference} (a). For single-step generation Fig.~\ref{inference} (b), we sample Gaussian noise $\boldsymbol{\epsilon}\sim\mathcal{N}(\mathbf{0},\mathbf{I})$, project it onto the latent sphere by spherification, and decode the resulting latent representation conditioned on the text embedding to obtain the initial image $\mathbf{x}^{(1)}$.

Generation quality can be further improved using iterative encoder--decoder refinement, as illustrated in Fig.~\ref{inference} (c). Specifically, given the initial image $x^{(1)}$ from single-step generation, the text-conditioned image encoder maps it to the image latent $z$. The latent is first projected onto the hypersphere via RMS normalization, followed by Gaussian perturbation $\sigma\epsilon^{(t)}$, where $\boldsymbol{\epsilon}\sim\mathcal{N}(\mathbf{0},\mathbf{I})$ and $\sigma$ controls the perturbation direction. This perturbation introduces stochastic variation, facilitating exploration of nearby regions in the latent space. Since the perturbation may displace the latent from the spherical manifold, a second spherification operation reprojects the perturbed image latent $\Bar{z}$ onto the hypersphere while preserving the noise-induced directional variation. This process is repeated for $T$ refinement steps, and the resulting latent is finally passed to the text-conditioned decoder to produce a refined image $x^{(t+1)}$.


%% file: sec/5_experiment.tex
\section{Experiments}
\label{Experiments}


\paragraph{Datasets.} We evaluate our framework on three established text-to-image benchmarks: CUB-200-2011~\cite{wah2011caltech}, MS-COCO-2014~\cite{lin2014microsoft}, and Flickr30k~\cite{young2014image}. Although larger datasets, such as CC3M/CC12M~\cite{sharma2018conceptual,changpinyo2021conceptual} and LAION~\cite{schuhmann2021laion,schuhmann2022laion}, are available, their storage requirements exceed our resources. The selected benchmarks nevertheless provide diverse evaluation settings, covering fine-grained objects, complex scenes, and varied text–image relationships.

CUB-200-2011 contains 11,788 images across 200 bird species, each with 10 captions, split into 8,855 training and 2,933 test images. MS-COCO-2014 contains approximately $82k$ training and $40k$ test images across 80 object categories, each with 5 captions, featuring complex multi-object scenes. Flickr30k comprises 31,783 images, each with 5 captions, of which 30,783 are used for training and 1,000 for testing, covering diverse everyday scenes.



\paragraph{Evaluation Metrics.} We evaluate reconstruction performance using reconstruction FID (rFID) and Inception Score (rIS), and inference/generation performance using generation FID (gFID), generation IS (gIS), and CLIPScore. Fréchet Inception Distance (FID)~\cite{heusel2017gans} measures the distributional discrepancy between real and generated images, while Inception Score (IS)~\cite{salimans2016improved} assesses sample quality and diversity using a pretrained Inception network. CLIPScore~\cite{hessel2021clipscore} measures text–image semantic consistency through the cosine similarity of CLIP embeddings. \textbf{We adopt gFID as the primary metric.} Unlike gIS, which relies on classification-based predictions, gFID explicitly compares generated images against real data. Reconstruction metrics, in contrast, assess the model's ability to recover existing images rather than its generative performance. Our experiments further reveal that improved reconstruction fidelity does not necessarily translate into better generation quality, as demonstrated in the ablation study in Section~\ref{reconstruction_generation_analysis}. Moreover, CLIPScore provides only an approximate measure of text–image semantic alignment. Therefore, we report rFID, rIS, gIS, and CLIPScore as complementary metrics to provide a more comprehensive evaluation of reconstruction quality, generation performance, and semantic alignment.



\paragraph{Experimental Setup.} Input images are resized to $256 \times 256$, with a patch size of $16 \times 16$. We retain the original ViT-Large encoder and decoder architectures of Sphere Encoder, with a latent token dimension of 16 and 4 latent MLP-Mixer layers. For text conditioning, we use a frozen CLIP ViT-L/14@336px text encoder with a maximum token length of 77.

For the proposed SSAU objective, we set $\lambda_{\mathrm{SSAU}} = 1.0$ to maintain a weighting consistent with the original Sphere Encoder objective, while retaining all hyperparameters of $\mathcal{L}_{\mathrm{Sphere}}$. The uniformity weight $\lambda_u = 0.2$ is determined through an ablation study 
(see Table~\ref{reconstruction_metrics_weights} in the Supplement). We train for 600 epochs on CUB-200-2011 and 100 epochs on each of MS-COCO-2014 and Flickr30k. 

All training and evaluation experiments are conducted using PyTorch on 4 NVIDIA GH200 Grace Hopper GPUs. Training requires approximately 1, 3, and 1.5 GPU-days for CUB-200-2011, MS-COCO-2014, and Flickr30k, respectively. For evaluation, we generate 50K images at $256 \times 256$ resolution per dataset, requiring approximately 6 GPU-hours per dataset. Additionally, we measure inference efficiency on a single NVIDIA RTX 2080Ti GPU, with an average latency of 7 seconds per image per refinement step.




\section{Results and Analysis}
\label{Statistical Results}

\begin{table}[t]
\centering
\caption{Quantitative reconstruction performance across three datasets.}
\label{reconstruction_metrics}
\begin{tabular}{cc|cc|cc}
\toprule
\multicolumn{2}{c|}{CUB-200-2011} &
\multicolumn{2}{c|}{MS-COCO-2014} &
\multicolumn{2}{c}{Flickr30k} \\
\cmidrule(lr){1-2}
\cmidrule(lr){3-4}
\cmidrule(lr){5-6}
rFID $\downarrow$ & rIS $\uparrow$ &
rFID $\downarrow$ & rIS $\uparrow$ &
rFID $\downarrow$ & rIS $\uparrow$ \\
\midrule
0.92 & 5.86 & 0.78 & 36.25 & 0.82 & 24.17 \\
\bottomrule
\end{tabular}
\end{table}



\paragraph{Evaluation of Image Reconstruction.} We begin by evaluating the reconstruction capability of our framework across three datasets to assess its ability to preserve visual information in the spherical latent space. As reported in Table~\ref{reconstruction_metrics}, we evaluate 50K reconstructed images per dataset using reconstruction FID (rFID) and Inception Score (rIS). Our framework achieves rFID below 1.0 across all three datasets, indicating strong reconstruction fidelity. In terms of rIS, MS-COCO-2014 achieves the highest score of 36.25, followed by Flickr30k (24.17) and CUB-200-2011 (5.86). Qualitative results in Fig.~\ref{reconstruction} of the Supplement further demonstrate faithful reconstructions of fine-grained bird species in CUB-200-2011 and complex real-world scenes in MS-COCO-2014 and Flickr30k.

\begin{table}[t]
\centering
\caption{Best-performing configurations for each dataset in terms of gFID. More details are provided in Fig.~\ref{cfg_step} in the Supplement.}
\label{generation_metrics}
\footnotesize
\setlength{\tabcolsep}{3.5pt}
\begin{tabular}{lccccc}
\toprule
Dataset & CFG & Steps & gFID $\downarrow$ & gIS $\uparrow$
& CLIPScore $\uparrow$ \\
\midrule
CUB-200-2011 & 9.0 & 16 & 6.82 & 5.34 & 24.74 \\
MS-COCO-2014 & 2.0 & 6 &  7.11 & 28.67 & 24.11 \\
Flickr30k & 3.0  & 8 & 13.86 & 17.20 & 23.65 \\
\bottomrule
\end{tabular}
\end{table}


\paragraph{Evaluation of Image Generation.}
We evaluate the text-to-image generation performance of our framework across three datasets, investigating the effects of
classifier-free guidance (CFG) and iterative encoder--decoder refinement. Following the original Sphere Encoder, we employ CFG to control the strength of textual conditioning during inference. We first vary the CFG scale from 2.0 to 10.0 in increments of 1.0, while fixing the number of forward steps at 4 based on the reliable performance reported by the original Sphere Encoder. Using gFID as the primary evaluation criterion, we select the best-performing CFG scale for each dataset: 9.0 for CUB-200-2011, 2.0 for MS-COCO-2014, and 3.0 for Flickr30k.
Detailed results and analysis are presented in the left panel of Fig.~\ref{cfg_step} in the Supplement. 

Using the selected dataset-specific CFG scales, we further investigate the effect of iterative refinement by varying the number of forward steps from 4 to 16 in increments of 2.
Generation performance is evaluated using gFID, gIS, and CLIPScore, with detailed results presented in the right panel of Fig.~\ref{cfg_step} in the Supplement. 
Table~\ref{generation_metrics} summarizes the best-performing configurations for each dataset in terms of gFID. Specifically, CUB-200-2011 achieves the lowest gFID of 6.82 at 16 steps, while MS-COCO-2014 and Flickr30k achieve their lowest gFID values of 7.11 and 6 at 13.86 and 8 steps, respectively. These results demonstrate the effectiveness of our framework in generating high-quality images with a limited number of refinement steps. Notably, increasing the number of steps does not consistently improve generation
performance, suggesting that excessive refinement is not always beneficial and that the optimal step count varies across datasets.


Qualitatively, Fig.~\ref{combine_sample_at_each_step} illustrates the progressive generation process over the first 8 refinement steps. Our framework captures the primary semantic content of the input captions in a single step, generating recognisable objects and scene structures, including a black-feathered bird on CUB-200-2011, an airborne snowboarder on MS-COCO-2014, and people around picnic tables on Flickr30k. Subsequent refinement progressively improves object boundaries, structural coherence, and visual details,
with substantial improvements observed within the first 4 steps. These observations further highlight the ability of our framework to establish semantic content early and efficiently refine visual fidelity through
iterative generation.

\begin{table*}[t]
\centering
\caption{Comparison with diffusion-based methods trained from scratch.}
\label{diffusion_comparison}
\begin{tabular}{ccccccc}
\toprule
\multirow{2}{*}{Methods} &
\multicolumn{3}{c}{CUB-200-2011} &
\multicolumn{3}{c}{MS-COCO-2014} \\
\cmidrule(lr){2-4} \cmidrule(lr){5-7}
& Steps & gFID $\downarrow$ & gIS $\uparrow$
& Steps & gFID $\downarrow$ & gIS $\uparrow$ \\
\midrule
Improved VQ-Diffusion~\cite{tang2022improved}
& 100 & 10.21 & -- & 100 & 8.44 & -- \\
VQ-Diffusion~\cite{gu2022vector}
& 100 & 10.32 & -- & 100 & 13.86 & -- \\
TCTS+FAS~\cite{lee2023text}
& 16 & 12.35 & -- & 16 & 14.45 & -- \\
UNIFIED DISCRETE~\cite{hu2022unified}
& 500 & 16.19 & \textbf{6.02} & 500 & 25.11 & 21.03 \\
Ours
& \textbf{4/16} & \textbf{8.03/6.82} &  5.38/5.34 & \textbf{6}
& \textbf{7.11} & \textbf{28.67} \\
\bottomrule
\end{tabular}
\end{table*}

\begin{figure}[!t]
\centering
\includegraphics[width=8.3cm]{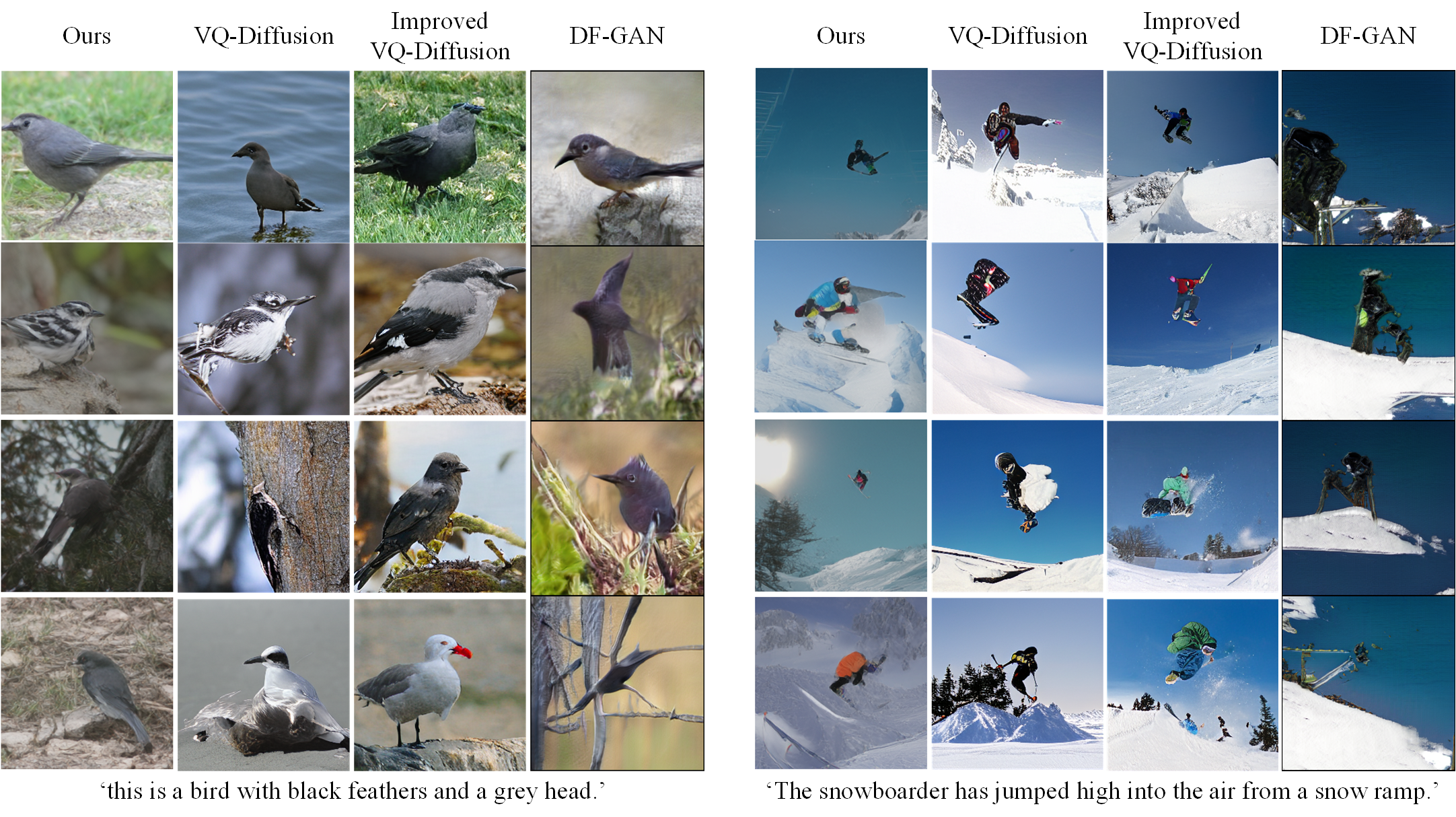}
\caption{Qualitative comparison with prior methods using publicly available checkpoints on CUB-200-2011 (left) and MS-COCO-2014 (right).}
\label{vision_compare}
\end{figure}


\begin{table}[t]
\centering
\caption{Comparison with GAN-based methods trained from scratch.}
\label{gan_comparison}
\begin{tabular}{ccccc}
\toprule
\multirow{2}{*}{Methods} &
\multicolumn{2}{c}{CUB-200-2011} &
\multicolumn{2}{c}{MSCOCO2014} \\
\cmidrule(lr){2-3} \cmidrule(lr){4-5}
& gFID $\downarrow$ & gIS $\uparrow$
& gFID $\downarrow$ & gIS $\uparrow$ \\
\midrule
HyperCGAN~\cite{haydarov2024adversarial}  & 11.00 & -- & 18.58 & -- \\
ALR-GAN~\cite{tan2023alr}  & 15.14 & 4.96 & 29.04 & 34.70 \\
DFGAN~\cite{tao2022df}  & 14.81 & 5.10 & 19.32 & -- \\
LAFITE~\cite{zhou2022towards}  & 10.48 & \textbf{5.97} & 8.12 & 32.34 \\
RAT-GAN~\cite{ye2023recurrent}  & 13.91 & 5.36 & 14.60 & -- \\
DSE-GAN~\cite{huang2022dse} & 13.23 & 5.13 & 15.30 & 26.71 \\
DR-GAN~\cite{tan2022dr} & 14.96 & 4.90 & 27.80 & 34.59 \\
DAE-GAN~\cite{ruan2021dae} & 15.19 & 4.42 & 28.12 & \textbf{35.08} \\
SSA-GAN~\cite{liao2022text} & 15.61 & 5.17 & 19.37 & -- \\
TIME~\cite{liu2021time} & 14.30 & 4.91 & 31.14 & 30.85 \\
TVBi-GAN~\cite{wang2020text} & 11.83 & 5.03 & 31.97 & 31.01 \\
KT-GAN~\cite{tan2020kt} & 17.32 & 4.85 & 30.73 & 31.67 \\
SEGAN~\cite{tan2019semantics} & 18.16 & 4.67 & 32.27 & 27.86 \\
DM-GAN~\cite{zhu2019dm} & 16.09 & 4.75 & 32.64 & 30.49 \\
DAI-TIS~\cite{lao2019dual} & 18.41 & 3.58 & 27.07 & 8.94 \\
StackGAN++~\cite{zhang2018stackgan++} & 15.30 & 4.04 & 81.59 & 8.30 \\
Ours & \textbf{6.82} & 5.34 & \textbf{7.11} & 28.67 \\
\bottomrule
\end{tabular}
\end{table}

\paragraph{Comparison with Existing Methods.}
We compare our framework with diffusion-based and GAN-based methods
on CUB-200-2011 and MS-COCO-2014 in
Tables~\ref{diffusion_comparison} and~\ref{gan_comparison},
respectively, as existing methods do not report Flickr30k results. On CUB-200-2011, our framework achieves gFID values of 8.03 and 6.82 with only 4 and 16 inference steps, respectively, outperforming diffusion-based methods that require over 16 steps, while maintaining a competitive gIS score of 5.3. On MS-COCO-2014, it achieves the lowest gFID of 7.11 and highest gIS of 28.67 with only 6 steps. Compared with GAN-based methods, our framework achieves the lowest gFID on both CUB-200-2011 (6.82) and MS-COCO-2014 (7.11), although LAFITE~\cite{zhou2022towards} achieves higher gIS. These results demonstrate strong distributional fidelity with substantially fewer inference steps than diffusion-based counterparts.

For qualitative comparison, we evaluate our framework against VQ-Diffusion~\cite{gu2022vector}, Improved VQ-Diffusion~\cite{tang2022improved}, and DF-GAN~\cite{tao2022df}, using publicly available checkpoints and identical captions. As shown in Fig.~\ref{vision_compare}, our framework generates visually coherent images with accurate semantic details, including bird appearances and snowboarder poses. In contrast, competing methods exhibit inconsistencies in object structures or caption-specific attributes. Notably, our framework requires only 16 inference steps, compared with 100 steps for the diffusion-based methods, demonstrating its effectiveness in efficient text-to-image generation.

\begin{figure*}[t]
\centering
\includegraphics[width=16cm]{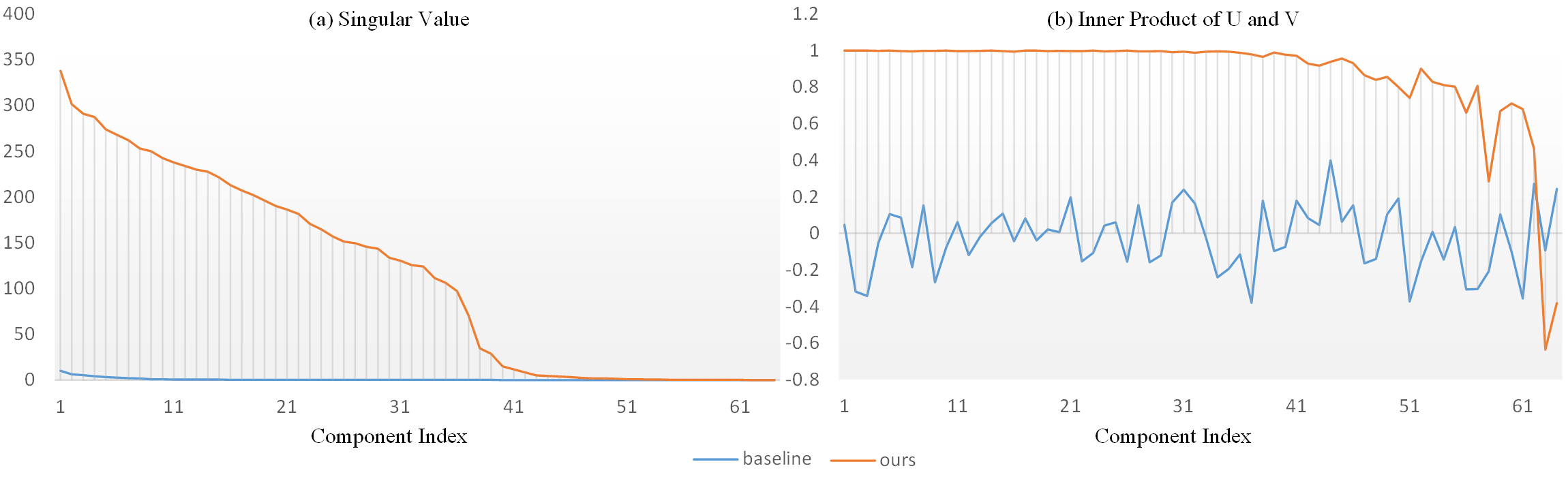}
\caption{PLS-SVD analysis of the text-conditioned Sphere Encoder with the proposed alignment loss $\mathcal{L}_{\mathrm{SSAU}}$ on the CUB-200-2011 dataset.}
\label{figure2_pls_svd}
\end{figure*}

\subsection{Ablation Study}
\label{ablation_study_main_page}

We conduct ablation studies on CUB-200-2011, MS-COCO-2014, and Flickr30k to investigate the contribution of explicit text-image alignment and the sensitivity to uniformity regularization. Performance is assessed using rFID, rIS, gFID, gIS, and CLIPScore, with gFID as the primary metric.

We first evaluate the contribution of $\mathcal{L}_{\mathrm{SSAU}}$ by removing it while keeping all other settings unchanged, examining its impact on reconstruction and generation performance (see Table~\ref{contribution_align_loss_cub}, ~\ref{contribution_align_loss_coco}, ~\ref{contribution_align_loss_flickr} in the Supplement). We further investigate the influence of uniformity regularization by comparing $\lambda_u = 0.1$ and $0.2$, with the latter achieving better overall performance (see Table~\ref{reconstruction_metrics_weights} and Fig.~\ref{cfg_step_weights} in the Supplement). These results highlight the importance of explicit cross-modal alignment and support the selected uniformity weight. Detailed results are provided in the Supplement. 



\subsection{Cross-Modal Representation Analysis}
\label{pls_svd_analysis}



To investigate the effectiveness of our proposed SSAU objective, we employ PLS-SVD to analyze the cross-covariance structure between text and image representations. We compare our framework against the baseline Sphere Encoder adapted to text conditioning without explicit cross-modal alignment.

As shown in Fig.~\ref{figure2_pls_svd}, the maximum singular value $\Sigma_{ii}$ increases from approximately 10 to 350,
with substantial singular values spanning the first 40 components, compared with 10 in the baseline. Furthermore, the $U$-$V$ alignment scores remain close to 1.0 across the first 40 components and above 0.3 within the first 60 components,
whereas the baseline exhibits weaker and less stable
correspondence, fluctuating between -0.3 and 0.3.
These findings suggest that SSAU strengthens cross-modal
covariance and directional correspondence, encouraging
more coherent representation structures between text
and image modalities.




\subsection{Reconstruction \textit{vs.} Generation Performance}
\label{reconstruction_generation_analysis}

Our ablation study in Supplement 
reveals a discrepancy between reconstruction
and generation performance. Specifically, removing the proposed
SSAU objective improves reconstruction fidelity, yielding a
lower rFID, but degrades generation quality, resulting in a
higher gFID. This suggests that improved reconstruction fidelity does not necessarily translate into better generation performance. While reconstruction primarily measures the ability to preserve visual information from existing images, generation additionally requires learning representations that support realistic text-conditioned sampling. These findings highlight the importance of explicit cross-modal regularization and further justify our use of gFID as the primary evaluation metric.

\subsection{Limitations}
\label{limitations}

Despite improvements in generation quality and cross-modal alignment, our framework has several limitations. First, generation performance remains sensitive to CFG scales and refinement steps, requiring dataset-specific configurations. Second, our experiments are limited to CUB-200-2011, MS-COCO-2014, and Flickr30k, without evaluation on larger-scale datasets, leaving the scalability and generalization of our approach to more diverse and complex data unexplored. 

%% file: sec/6_conclusion.tex

\section{Conclusion}


In this work, we introduce \texttt{TextSphere} to extend the Sphere Encoder for efficient text-to-image generation and propose an \textbf{SSAU} objective to strengthen cross-modal correspondence in the spherical latent space. By combining symmetric contrastive learning with spherical uniformity regularization, our approach improves text-image alignment while preserving generation quality without modifying the original image encoder and decoder architectures. Experiments on CUB-200-2011, MS-COCO-2014, and Flickr30k demonstrate competitive performance with substantially fewer inference steps than existing diffusion-based methods. Extensive ablation studies validate SSAU's effectiveness and show that improved reconstruction fidelity does not necessarily translate into better generation quality. Furthermore, PLS-SVD analysis provides representation-level evidence of stronger cross-modal covariance and directional correspondence. These findings highlight the importance of explicitly aligning multimodal representations for efficient text-to-image generation in the framework. Future work will investigate scalability to larger datasets and more robust inference strategies.

%% file: sec/X_suppl.tex
\clearpage
\setcounter{page}{1}
\maketitlesupplementary

\section*{Related Work}
\label{related_work}

Text-to-image generation aims to synthesize visually realistic images that are semantically consistent with natural-language descriptions. Early approaches predominantly relied on Generative Adversarial Networks (GANs). The first widely recognized GAN-based text-to-image generation model is GAN-INT-CLS~\cite{reed2016generative}, which combines adversarial learning with text conditioning, matching-aware discrimination, and text embedding interpolation to generate realistic images that are semantically aligned with textual descriptions. AttnGAN~\cite{xu2018attngan} improves upon earlier GAN-based methods by incorporating word-level attention and multi-stage image refinement, enabling the generation of more detailed images with stronger semantic alignment to textual descriptions. Subsequent methods, such as DM-GAN~\cite{zhu2019dm}, further reduce dependence on the initial image quality and improve visual realism and semantic consistency by dynamically selecting and integrating relevant textual information during refinement. To simplify generation, DF-GAN~\cite{tao2022df} adopted a single-stage architecture with deep text–image feature fusion. More recently, LAFITE~\cite{zhou2022towards} leveraged pretrained CLIP representations to facilitate text-to-image generation with reduced reliance on paired training data, while GALIP~\cite{tao2023galip} incorporated CLIP-based visual representations into adversarial learning to improve generation quality and efficiency. Despite their relatively efficient inference, GAN-based approaches can face challenges in training stability and capturing diverse, complex image distributions.

Autoregressive models have also demonstrated strong capabilities in text-to-image synthesis by representing images as sequences of discrete visual tokens. DALL-E~\cite{ramesh2021zero} employed a Transformer to autoregressively model textual and visual tokens, demonstrating the effectiveness of large-scale multimodal learning. Building upon this paradigm, Parti~\cite{yu2022scaling} further explored the scalability of autoregressive Transformers for high-fidelity text-conditioned image generation. Alternatively, Muse~\cite{chang2023muse} introduced masked-token prediction with parallel decoding to improve generation efficiency. These approaches achieve remarkable generation quality by modeling images as sequences of discrete visual tokens. However, their large-scale Transformer architectures and sequential token prediction introduce substantial computational and inference costs. Muse improves generation efficiency through parallel masked-token prediction, but still relies on large pretrained models and discrete image tokenization.

More recently, diffusion-based methods have become a prominent paradigm for text-to-image generation because they can model complex image distributions and produce high-quality samples. GLIDE~\cite{nichol2021glide} demonstrates the effectiveness of diffusion models in generating photorealistic and semantically consistent images. However, its iterative denoising process requires multiple inference steps, resulting in substantial computational overhead. Similarly, Imagen~\cite{saharia2022photorealistic} improves image quality and text–image alignment by integrating powerful pretrained language representations with cascaded diffusion models. Nevertheless, its reliance on a large text encoder and multiple denoising stages further increases computational complexity and inference costs. To reduce the computational demands of pixel-space diffusion, Latent Diffusion Models (LDMs)~\cite{rombach2022high} perform denoising in a compressed latent space while incorporating textual conditions through cross-attention. VQ-Diffusion~\cite{gu2022vector} introduced discrete diffusion over vector-quantized image representations, while Improved VQ-Diffusion~\cite{tang2022improved} further enhanced generation quality through improved sampling and classifier-free guidance. More recently, diffusion Transformer-based approaches, including PixArt-$\alpha$~\cite{chen2024pixart}, have shown that Transformer architectures can enable scalable, high-resolution text-to-image generation. Nevertheless, conventional diffusion-based methods typically require multiple iterative sampling steps to progressively transform noise into realistic images.

To address the computational cost of iterative generation, recent studies have explored single-step and few-step image synthesis. Latent Consistency Models (LCMs)~\cite{luo2023latent} enable high-quality generation with only a few inference steps by distilling the sampling trajectories of pretrained latent diffusion models. InstaFlow~\cite{liu2023instaflow} uses rectified flow and trajectory straightening to achieve one-step text-to-image generation, while Distribution Matching Distillation (DMD)~\cite{chen2026flash} reduces sampling to a single step by matching the generated distribution to that of a pretrained diffusion model. Although these methods significantly accelerate inference, they generally rely on pretrained generative models or teacher-based distillation.



\section*{Qualitative Results of Image Reconstruction}

\begin{figure}[h]
\centering
\includegraphics[width=8.3cm]{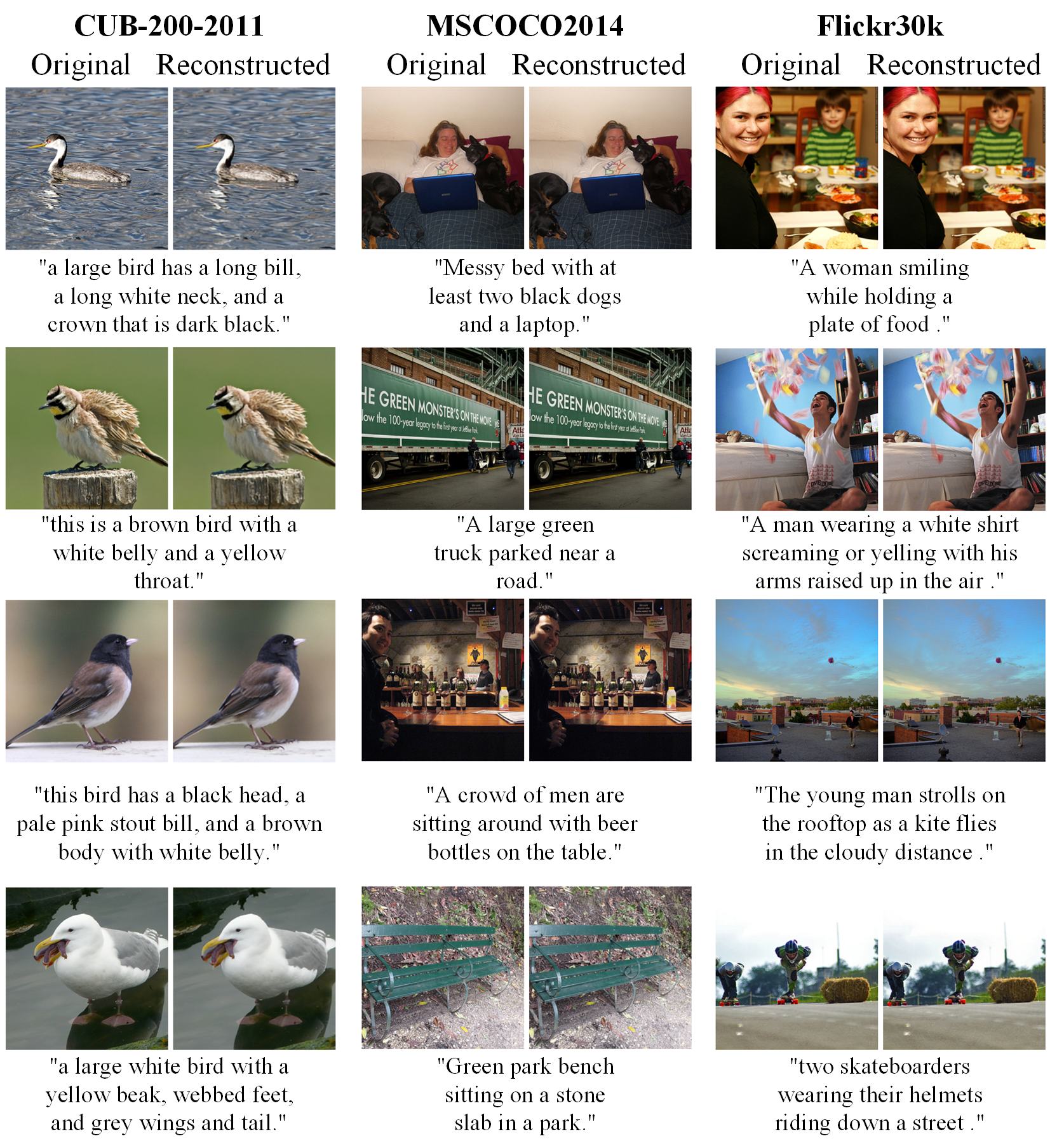}
\caption{Reconstructed images across three datasets.}
\label{reconstruction}
\end{figure}

To evaluate our framework's reconstruction capability, we project the encoded representations of input images onto the corresponding hypersphere and decode the resulting spherical latents to obtain reconstructed images across three datasets. Fig.~\ref{reconstruction} presents qualitative comparisons between the original images and their corresponding reconstructions. Notably, our framework consistently achieves high-fidelity reconstruction across diverse visual domains. On CUB-200-2011, the framework effectively captures distinctive bird characteristics, including variations in plumage color, beak shape, and body structure. On MS-COCO-2014, it successfully reconstructs diverse real-world scenes involving multiple objects, such as animals, vehicles, and outdoor environments, while preserving their spatial arrangements and contextual details. Similarly, on Flickr30k, the framework reconstructs human-centric scenes with diverse activities, body poses, and interactions with the surrounding environment. These results highlight the robustness and generalisability of our framework across datasets with varying visual complexity and semantic content.


\section*{Effect of CFG Scale and Refinement Steps}

The original Sphere Encoder framework already supports classifier-free guidance (CFG) to improve generation quality by controlling the strength of conditional information during inference. Building upon this capability, we further investigate the effectiveness of CFG in our \texttt{TextSphere} framework across three datasets. Specifically, we first evaluate how CFG scale affects generation performance by varying it from 2.0 to 10.0 in 1.0 increments. We fix the number of forward steps at 4 for this experiment. Next, we examine the effect of refinement steps. Since gFID is our primary evaluation criterion, we select the CFG scale that yields the lowest gFID for each dataset and use these dataset-specific settings in the subsequent analysis of iterative encoder–decoder refinement.

\begin{figure*}[h]
\centering
\includegraphics[width=13cm]{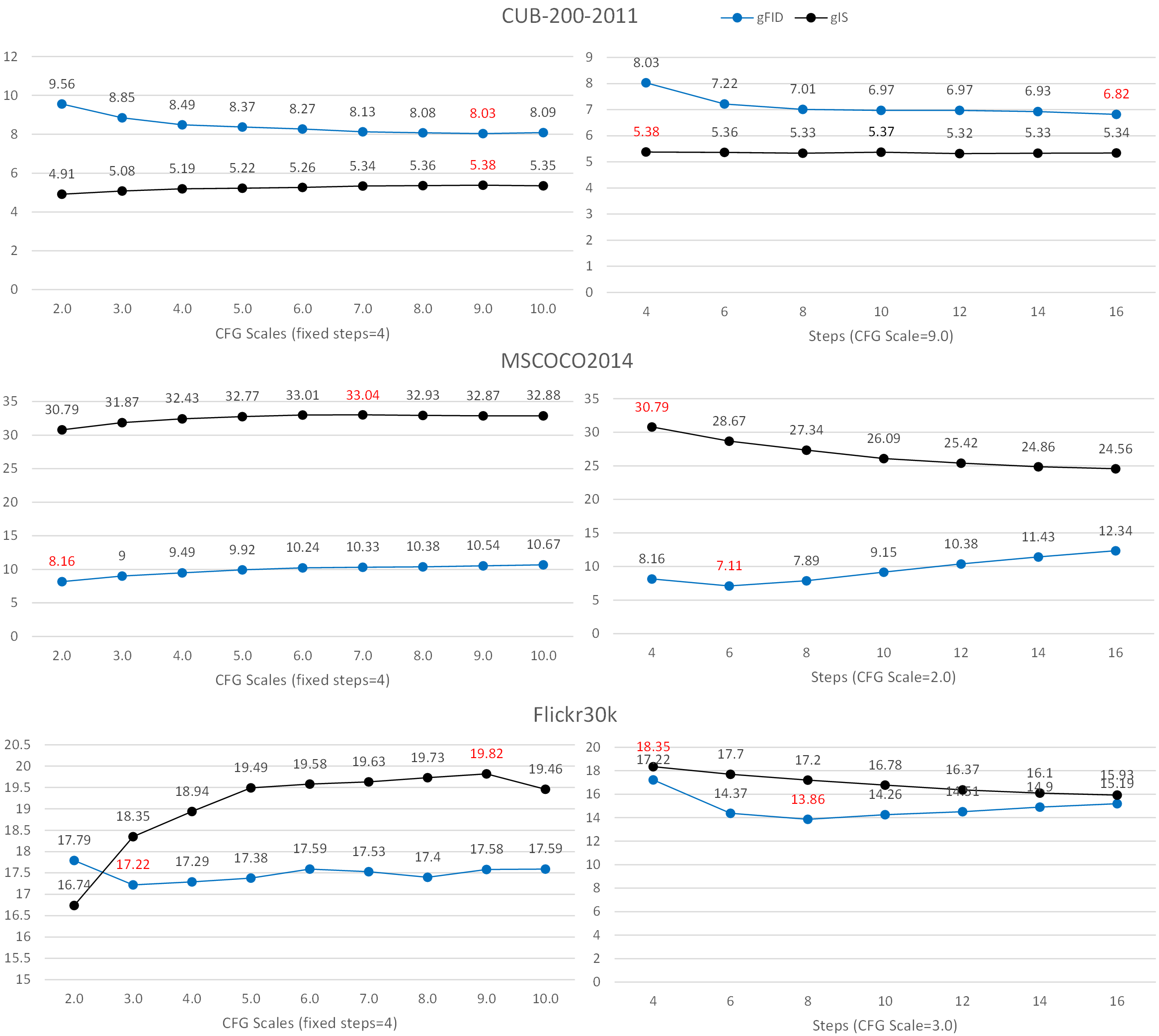}
\caption{Effect of CFG scale and iterative refinement steps on text-conditioned Sphere Encoder across three datasets. Left: gFID ($\downarrow$) and gIS ($\uparrow$) under varying CFG scales with the number of forward steps fixed at 4. Right: gFID and gIS across different refinement steps (4$-$16, interval 2), using the dataset-specific CFG scales that achieve the lowest gFID, \textit{i.e.}, best CFG=9.0 on CUB, CFG=2.0 on COCO, CFG=3.0 on Flickr30k.}
\label{cfg_step}
\end{figure*}

\paragraph{Effect of CFG Scale.} The left chart of Fig.~\ref{cfg_step} shows how gFID and gIS change across different classifier-free guidance (CFG) scales, with the number of forward steps fixed at 4. On CUB-200-2011, increasing the CFG scale consistently improves generation performance, with gFID decreasing from 9.56 to 8.03 and gIS increasing from 4.91 to 5.38 as the CFG scale reaches 9.0. These improvements indicate that stronger textual conditioning enhances both the distributional fidelity and visual quality of generated images, demonstrating the effectiveness of CFG in our text-conditioned Sphere Encoder.

In contrast, MS-COCO-2014 exhibits a different trend. The lowest gFID of 8.16 is achieved at a CFG scale of 2.0, beyond which gFID progressively increases to 10.67 as the guidance strength increases. Meanwhile, gIS continues to improve from 30.79 to 33.04 until the CFG scale reaches 7.0, followed by a slight decline at higher guidance scales. These observations suggest that excessively strong textual guidance does not necessarily translate into better generation quality. Although moderate increases in CFG scale may improve the visual characteristics captured by gIS, they can also increase distributional discrepancies between real and generated images, as reflected in the rising gFID. We hypothesize that this behavior is associated with the greater visual and semantic complexity of MS-COCO-2014, which contains diverse objects, complex backgrounds, and intricate spatial relationships. Compared with the relatively homogeneous bird images in CUB-200-2011, generating realistic images from complex textual descriptions requires balancing semantic consistency with visual fidelity. Excessively strong guidance may overemphasize textual conditioning, potentially restricting the flexibility of the generative process and shifting the generated image distribution away from the real data distribution.

A similar trend is observed on Flickr30k, where gFID initially decreases from 17.79 to 17.22 as the CFG scale increases from 3.0 to 10.0, but subsequently exhibits a gradual increase at higher guidance scales. In contrast, gIS continues to improve from 16.74 to 19.82 until the CFG scale reaches 9.0, suggesting that stronger textual conditioning can improve the visual characteristics measured by gIS without necessarily reducing the distributional discrepancy measured by gFID. Similar to MS-COCO-2014, Flickr30k contains diverse real-world scenes and complex interactions among multiple objects. We hypothesize that excessive guidance may amplify specific semantic features while reducing the natural variation and visual realism required to match the underlying image distribution, thereby limiting further improvements in gFID.

Overall, these results reveal that the effectiveness of CFG is dataset-dependent. While stronger guidance benefits generation performance on the relatively homogeneous CUB-200-2011 dataset, MSCOCO2014 and Flickr30k exhibit a more pronounced trade-off between guidance strength and distributional fidelity. Furthermore, the divergence between gFID and gIS shows that improvements in Inception Score do not necessarily mean better alignment with the real image distribution.


\paragraph{Effect of Refinement Steps.} Based on the preceding CFG scale experiments, we obtain the dataset-specific CFG scales that achieve the lowest gFID, \textit{i.e.}, CFG = 9.0 for CUB-200-2011, CFG = 2.0 for MSCOCO2014, and CFG = 3.0 for Flickr30k. Using these optimal CFG scales, we further investigate the effect of iterative encoder–decoder refinement by varying the number of forward steps from 4 to 16 with an interval of 2. The corresponding gFID and gIS results across the three datasets are presented in the right chart of Fig.~\ref{cfg_step}.

On CUB-200-2011, gFID consistently decreases as the number of refinement steps increases, dropping from 8.03 at step 4 to 6.82 at step 16. Meanwhile, gIS remains relatively stable at approximately 5.30 across different refinement steps. These results indicate that iterative refinement progressively reduces the distributional discrepancy between generated and real images, while maintaining comparable visual quality and diversity as measured by gIS.

In contrast, MS-COCO-2014 shows a non-monotonic trend, with gFID initially decreasing from 8.16 at step 4 to a minimum of 7.11 at step 6, then gradually increasing as more refinement steps are added. Meanwhile, gIS consistently decreases from 30.79 to 24.56 as the number of steps increases. These observations suggest that, unlike the progressive denoising process commonly employed in diffusion models, iterative encoder–decoder refinement in our text-conditioned Sphere Encoder does not necessarily yield continuous improvements in generation quality. Instead, the refinement process repeatedly encodes generated images into spherical latent representations, introduces stochastic perturbations, and decodes the resulting latents to produce updated images. While moderate refinement may improve visual fidelity and structural coherence, excessive iterations may accumulate representation shifts or introduce unnecessary variations, potentially moving the generated image distribution away from the real data distribution. Furthermore, the declining gIS indicates that additional refinement does not necessarily improve the visual quality and diversity captured by this metric.

A similar trend is observed on Flickr30k, where gFID initially decreases from 17.22 at step 4 to 13.86 at step 8, before increasing to 15.19 at step 16. This suggests that generation quality benefits from a limited number of refinement steps, beyond which further refinement may become less effective or even detrimental. Meanwhile, gIS exhibits a consistent downward trend, decreasing from 18.35 at step 4 to 15.93 at the maximum evaluated step 16, following a pattern similar to that observed on MS-COCO-2014.

Overall, these results show that iterative encoder–decoder refinement is dataset-dependent and that more refinement steps do not necessarily improve generation performance. While CUB-200-2011 benefits from additional refinement, MS-COCO-2014 and Flickr30k achieve their lowest gFID values at intermediate steps, suggesting that excessive refinement may introduce diminishing or negative returns. These findings highlight a fundamental distinction between spherical latent refinement and conventional diffusion-based denoising, where additional refinement steps are not inherently associated with improved image quality. Importantly, our framework achieves its best observed generation performance within a relatively small number of forward steps, demonstrating the potential for efficient text-to-image generation without requiring extensive iterative refinement. Accordingly, we select the number of forward steps achieving the lowest gFID for each dataset as its preferred inference configuration.

\subsection*{Ablation Study on Uniformity Weight $\lambda_u$}
\label{uniformity_weight_ablation}

\begin{figure*}[h]
\centering
\includegraphics[width=15cm]{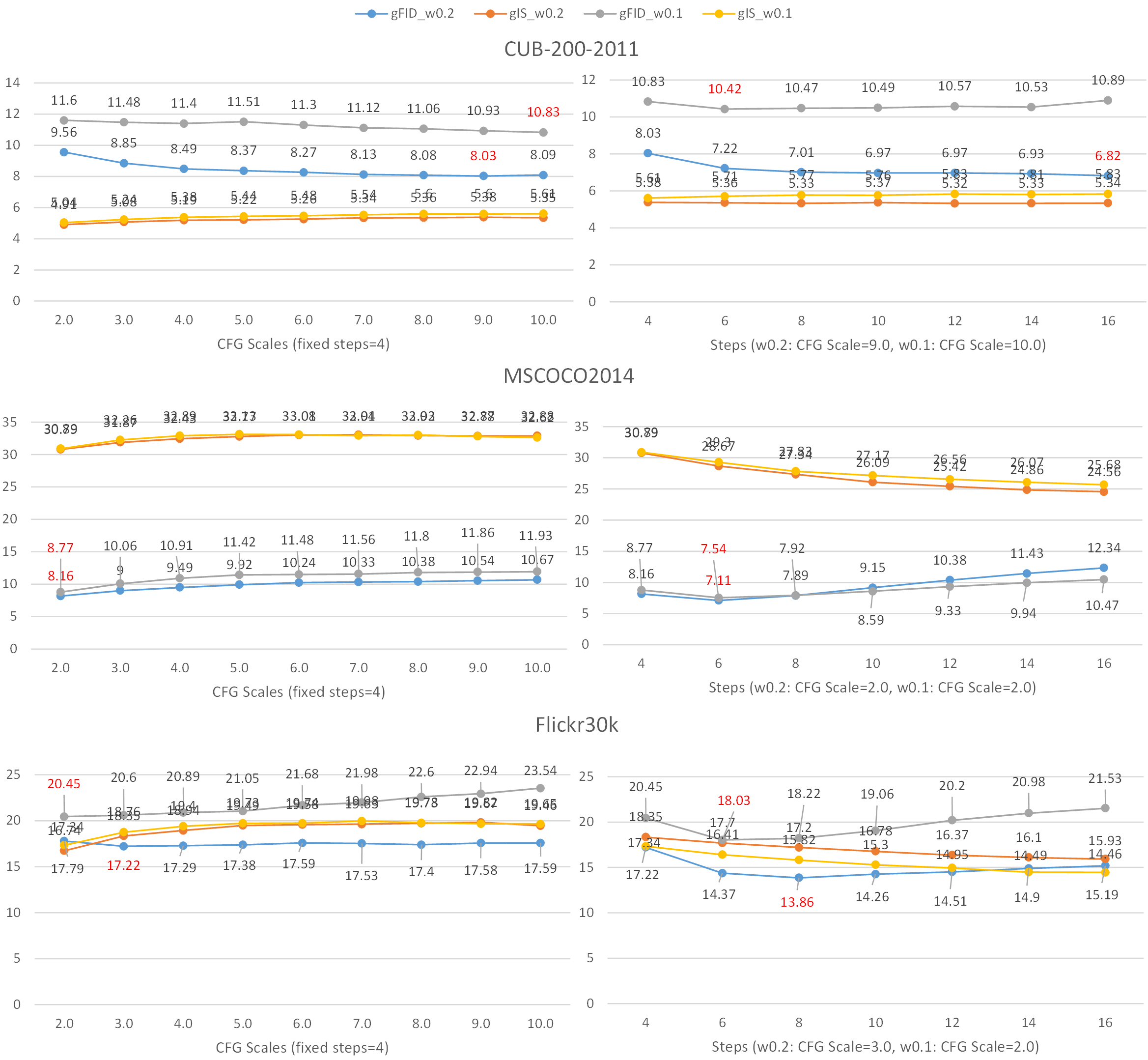}
\caption{Generation results of gFID and gIS using different uniformity weights across three datasets. w0.2 and w0.1 use uniformity weights of 0.2 and 0.1, respectively. Left: gFID and gIS comparisons under varying CFG scales with the number of forward steps at 4. Right: gFID and gIS across different refinement steps (4-16, interval 2), using the dataset-specific CFG scales that achieve the lowest gFID, \textit{i.e.}, best CFG=9.0 on CUB, CFG=2.0 on COCO, CFG=3.0 on Flickr30K.}
\label{cfg_step_weights}
\end{figure*}

To investigate the effect of the alignment uniformity weight $\lambda_u$ on model performance across three datasets, we evaluate two values, 0.1 and 0.2, with the corresponding results reported in Table~\ref{reconstruction_metrics_weights} and Fig.~\ref{cfg_step_weights}. 

As shown in Table~\ref{reconstruction_metrics_weights}, setting $\lambda_u = 0.2$ achieves lower rFID and higher rIS on the CUB-200-2011 and Flickr30k datasets compared to $\lambda_u = 0.1$. On MS-COCO-2014, $\lambda_u = 0.2$ yields a lower rFID but a slightly lower rIS. These results indicate that $\lambda_u = 0.2$ yields better reconstruction performance.

\begin{table}[t]
\centering
\caption{Reconstruction results of rFID and rIS using different uniformity weights across three datasets.}
\label{reconstruction_metrics_weights}
\begin{tabular}{c|cc|cc|cc}
\toprule
\multirow{2}{*}{$\lambda_{\mathrm{u}}$} & \multicolumn{2}{c|}{CUB} &
\multicolumn{2}{c|}{COCO} &
\multicolumn{2}{c}{Flickr} \\
\cmidrule(lr){2-3}
\cmidrule(lr){4-5}
\cmidrule(lr){6-7}
& rFID & rIS &
rFID & rIS &
rFID & rIS \\
\midrule
0.2 & \textbf{0.92} & \textbf{5.86} & \textbf{0.78} & 36.25 & \textbf{0.82} & \textbf{24.17} \\
0.1 & 2.11 & 5.80 & 0.89 & \textbf{36.29} & 0.98 & 24.12 \\
\bottomrule
\end{tabular}
\end{table}

To further examine the influence of classifier-free guidance (CFG) scales and the number of forward steps under different uniformity weights, we first fix the number of forward steps at 4 and vary the CFG scale from 2.0 to 10.0 in increments of 0.1 to identify the optimal CFG scale for each weight on each dataset. We then use the corresponding optimal CFG scale and vary the number of forward steps from 4 to 16 in increments of 2. We present the results in Fig.~\ref{cfg_step_weights}. As illustrated, $\lambda_u = 0.2$ generally outperforms $\lambda_u = 0.1$ in terms of generation quality, with the blue curves (gFID, $\lambda_u = 0.2$) consistently lying below the grey curves (gFID, $\lambda_u = 0.1$) in most settings. Although the two configurations exhibit comparable performance at higher numbers of forward steps on MS-COCO-2014, $\lambda_u = 0.2$ still achieves a lower minimum gFID of 7.11, compared to 7.54 for $\lambda_u = 0.1$.

Based on these observations, we set $\lambda_u = 0.2$ in our experiments, as a higher uniformity weight generally improves reconstruction and generation performance by more strongly encouraging separation of latent representations while promoting uniform coverage of the hyperspherical latent space.

\subsection*{Alation Study on AdaLN \textit{vs.} Cross-Attention}
\label{adaln_vs_ca}

In our preliminary experiments, we investigated replacing Adaptive Layer Normalization (AdaLN) with Cross-Attention (CA) for text--image conditioning within the original Sphere Encoder framework on the CUB-200-2011 dataset. Although CA is widely adopted in text-to-image synthesis for modeling fine-grained interactions between textual and visual representations, our empirical findings did not support our initial hypothesis, as CA consistently underperformed AdaLN within the Sphere Encoder architecture.

Following the original Sphere Encoder setup, we employ Classifier-Free Guidance (CFG)~\cite{ho2022classifier} during inference and evaluate guidance scales ranging from 2.0 to 10.0 in increments of 1.0. We initially fix the number of forward steps to 4, as the original Sphere Encoder demonstrates strong performance with four-step inference across multiple datasets. The quantitative results are presented in Table~\ref{ca_adaln_compare_cub}. Increasing the CFG scale generally improves the performance of both conditioning mechanisms. Specifically, gFID consistently decreases with stronger guidance, while gIS and CLIPScore exhibit overall increasing trends, suggesting improvements in generation quality, diversity, and text--image semantic alignment. Minor fluctuations are observed at higher guidance scales, where gIS and CLIPScore exhibit slight decreases in certain configurations. Nevertheless, a CFG scale of 10.0 achieves the strongest overall performance across the evaluated metrics.

To further investigate the influence of inference steps, we fix the CFG scale at 10.0 and vary the number of forward steps from 4 to 16 in increments of 2. The corresponding results are reported in the lower section of Table~\ref{ca_adaln_compare_cub}. Interestingly, the two conditioning mechanisms exhibit different trends as the number of forward steps increases. While CA generally benefits from additional steps, achieving progressively lower gFID values, AdaLN shows an increasing trend in gFID, indicating that additional refinement steps degrade generation quality. Despite this difference, AdaLN consistently achieves better overall generation performance than CA. Notably, the degradation observed with AdaLN suggests that increasing the number of inference steps alone does not necessarily improve generation quality in the text-conditioned Sphere Encoder without an explicit text--image alignment objective. In contrast, when we incorporate the proposed alignment loss $\mathcal{L}_{\mathrm{SSAU}}$, gFID generally decreases as the number of forward steps increases, suggesting that improved cross-modal alignment may let the model benefit more from iterative refinement. Regarding gIS and CLIPScore, both conditioning mechanisms exhibit relatively limited sensitivity to the number of forward steps, with AdaLN generally maintaining higher values than CA.

Overall, AdaLN outperforms CA across most evaluated CFG scales and forward-step configurations in terms of gFID, gIS, and CLIPScore. This advantage also extends to reconstruction performance, where AdaLN achieves a lower rFID and a higher rIS than CA, as reported in the upper section of Table~\ref{ca_adaln_compare_cub}. We hypothesize that this advantage partly stems from AdaLN's architectural compatibility with the interleaved RMS normalization layers in Sphere Encoder, which may promote more stable activation distributions and consistent feature scaling throughout the network, consistent with recent observations in~\cite{hacohen2026ltx}. These findings suggest that, despite the ability of CA to capture fine-grained cross-modal interactions, its additional representational flexibility does not necessarily translate into improved performance within the Sphere Encoder architecture. Accordingly, we retain AdaLN as the text-conditioning mechanism in all subsequent experiments.

\begin{table*}[t]
\centering
\caption{Quantitative comparisons using Cross-Attention (CA) \textit{vs.} AdaLN for text conditioning on CUB-200-2011 dataset. Top: reconstruction results. Middle: varying CFG scales with fixed forward steps=4. Bottom: varying forward steps with best CFG scale=10.0.}
\label{ca_adaln_compare_cub}
\begin{tabular}{c|cc|cc|cc}
\toprule
\multicolumn{7}{c}{\textit{Reconstruction Results (CFG/Step Independent)}} \\
\midrule
Metric & \multicolumn{3}{c|}{CA} & \multicolumn{3}{c}{AdaLN} \\
\midrule
rFID $\downarrow$ & \multicolumn{3}{c|}{1.76} & \multicolumn{3}{c}{\textbf{0.77}} \\
rIS $\uparrow$    & \multicolumn{3}{c|}{5.77} & \multicolumn{3}{c}{\textbf{5.89}} \\
\midrule
\multicolumn{7}{c}{\textit{Generation Results (fixed steps=4)}} \\
\midrule
\multirow{2}{*}{CFG} &
\multicolumn{2}{c|}{gFID $\downarrow$} &
\multicolumn{2}{c|}{gIS $\uparrow$} &
\multicolumn{2}{c}{CLIPScore $\uparrow$} \\
& CA & AdaLN & CA & AdaLN & CA & AdaLN \\
\midrule
2.0  & 15.29 & 10.67 & 4.40 & 4.77 & 24.44 & 24.67\\
3.0  & 14.24 & 10.36 & 4.53 & 4.87 & 24.65 & 25.03\\		
4.0  & 13.76 & 10.17 & 4.62 & 4.91 & 24.76 & 25.36\\			
5.0  & 13.49 & 10.11 & 4.64 & 4.92 & 24.87 & 25.59\\					
6.0  & 13.23 & 10.03 & 4.68 & 4.93 & 24.95 & 25.70\\					
7.0  & 12.80 & 9.80 & 4.76 & 4.98 & 25.03 & \textbf{25.76}\\				
8.0  & 12.47 & 9.70 & 4.76 & 4.97 & 25.12 & 25.74\\					
9.0  & 12.38 & 9.64 & 4.82 & 5.01 & 25.16 & 25.73\\					
10.0 & \textbf{12.13} & \textbf{9.42} & \textbf{4.83} & \textbf{5.02} & \textbf{25.24} & 25.67\\
\midrule
\multicolumn{7}{c}{\textit{Generation Results (CFG scale=10.0)}} \\
\midrule
\multirow{2}{*}{CFG} &
\multicolumn{2}{c|}{gFID $\downarrow$} &
\multicolumn{2}{c|}{gIS $\uparrow$} &
\multicolumn{2}{c}{CLIPScore $\uparrow$} \\
& CA & AdaLN & CA & AdaLN & CA & AdaLN \\
\midrule
4  & 12.13 & 9.42 & 4.83 & 5.02 & 25.24 & 25.67\\
6  & 10.94 & \textbf{9.10} & 4.84 & \textbf{5.04} & 25.36 & 25.80 \\		
8  & 10.54 & 9.41 & 4.84 & 5.00 & 25.42 & 25.79\\			
10  & 10.27 & 9.68 & 4.87 & 5.00 & 25.43 & 25.81\\		 			
12  & 10.31 & 10.01 & 4.87 & 5.00 & 25.45 & \textbf{25.83}\\				
14  & 10.29 & 10.07 & 4.84 & 4.97 & \textbf{25.46} & 25.83\\				
16  & \textbf{10.18} & 10.23 & \textbf{4.88} & 4.96 & 25.46 & 25.80\\		
\bottomrule
\end{tabular}
\end{table*}

\subsection*{Ablation Study on SSAU Objective $\mathcal{L}_{\mathrm{SSAU}}$}
\label{contribution_ssau}

To evaluate the contribution of the proposed SSAU loss $\mathcal{L}_{\mathrm{SSAU}}$, we compare our method against a baseline based on the original Sphere Encoder framework, in which the class embeddings are directly replaced with text embeddings extracted from a pretrained CLIP text encoder, without introducing the proposed SSAU loss. The quantitative reconstruction and generation results on CUB-200-2011, MS-COCO-2014, and Flickr30k are reported in Tables~\ref{contribution_align_loss_cub}, \ref{contribution_align_loss_coco}, and \ref{contribution_align_loss_flickr}, respectively.

As shown in Table~\ref{contribution_align_loss_cub}, our method exhibits slightly inferior reconstruction performance than the baseline on CUB-200-2011, with a marginally higher rFID and a slightly lower rIS. Similar trends are observed on MS-COCO-2014 and Flickr30k, as reported in Tables~\ref{contribution_align_loss_coco} and \ref{contribution_align_loss_flickr}. 

To further assess the impact of the proposed alignment loss on generation performance, we conduct an extensive evaluation across different classifier-free guidance (CFG) scales and numbers of forward steps. Specifically, we first fix the number of forward steps at 4 and vary the CFG scale to identify the optimal guidance configuration. On CUB-200-2011, our method achieves its lowest gFID of 8.03 and highest gIS of 5.38 at a CFG scale of 9.0. Using the selected CFG scale, we further increase the number of forward steps to 16, where our method achieves a minimum gFID of 6.82 and a gIS of 5.34. These results outperform the original baseline. Similar trends appear on MS-COCO-2014 and Flickr30k, as shown in Tables~\ref{contribution_align_loss_coco} and \ref{contribution_align_loss_flickr}, underscoring the importance of $\mathcal{L}_{\mathrm{Sphere}}$ in the proposed framework.

\begin{table*}[t]
\centering
\caption{Ablation study of the proposed SSAU objective on the CUB-200-2011 dataset.}
\label{contribution_align_loss_cub}
\begin{tabular}{c|cc|cc|cc}
\toprule
\multicolumn{7}{c}{\textit{Reconstruction Results (CFG/Step Independent)}} \\
\midrule
Metric & \multicolumn{3}{c|}{$\mathcal{L}_{\mathrm{Sphere}}+\lambda_{\mathrm{SSAU}}\mathcal{L}_{\mathrm{SSAU}}$ (Ours)} & \multicolumn{3}{c}{$\mathcal{L}_{\mathrm{Sphere}}$ (Original)} \\
\midrule
rFID $\downarrow$ & \multicolumn{3}{c|}{0.92} & \multicolumn{3}{c}{\textbf{0.77}} \\
rIS $\uparrow$    & \multicolumn{3}{c|}{5.86} & \multicolumn{3}{c}{\textbf{5.89}} \\
\midrule
\multicolumn{7}{c}{\textit{Generation Results (fixed steps=4)}} \\
\midrule
\multirow{2}{*}{CFG} &
\multicolumn{2}{c|}{gFID $\downarrow$} &
\multicolumn{2}{c|}{gIS $\uparrow$} &
\multicolumn{2}{c}{CLIPScore $\uparrow$} \\
& $\mathcal{L}_{\mathrm{Sphere}}+\lambda_{\mathrm{SSAU}}\mathcal{L}_{\mathrm{SSAU}}$ & $\mathcal{L}_{\mathrm{Sphere}}$ & $\mathcal{L}_{\mathrm{Sphere}}+\lambda_{\mathrm{SSAU}}\mathcal{L}_{\mathrm{SSAU}}$ & $\mathcal{L}_{\mathrm{Sphere}}$ & $\mathcal{L}_{\mathrm{Sphere}}+\lambda_{\mathrm{SSAU}}\mathcal{L}_{\mathrm{SSAU}}$ & $\mathcal{L}_{\mathrm{Sphere}}$ \\
\midrule
2.0  & 9.56 & 10.67 & 4.91 & 4.77 & 24.32 & 24.67\\
3.0  & 8.85 & 10.36 & 5.08 & 4.87 & 24.39 & 25.03\\		
4.0  & 8.49 & 10.17 & 5.19 & 4.91 & 24.47 & 25.36\\			
5.0  & 8.37 & 10.11 & 5.22 & 4.92 & 24.53 & 25.59\\					
6.0  & 8.27 & 10.03 & 5.26 & 4.93 & 24.58 & 25.70\\					
7.0  & 8.13 & 9.80 & 5.34 & 4.98 & 24.61 & \textbf{25.76}\\					
8.0  & 8.08 & 9.70 & 5.36 & 4.97 & 24.67 & 25.74\\					
9.0  & \textbf{8.03} & 9.64 & \textbf{5.38} & 5.01 & 24.71 & 25.73\\					
10.0 & 8.09 & \textbf{9.42} & 5.35 & \textbf{5.02} & \textbf{24.75} & 25.67\\
\midrule
\multicolumn{7}{c}{\textit{Generation Results (Ours: CFG scale=9.0. Original: CFG scale=10.0)}} \\
\midrule
\multirow{2}{*}{CFG} &
\multicolumn{2}{c|}{gFID $\downarrow$} &
\multicolumn{2}{c|}{gIS $\uparrow$} &
\multicolumn{2}{c}{CLIPScore $\uparrow$} \\
& $\mathcal{L}_{\mathrm{Sphere}}+\lambda_{\mathrm{SSAU}}\mathcal{L}_{\mathrm{SSAU}}$ & $\mathcal{L}_{\mathrm{Sphere}}$ & $\mathcal{L}_{\mathrm{Sphere}}+\lambda_{\mathrm{SSAU}}\mathcal{L}_{\mathrm{SSAU}}$ & $\mathcal{L}_{\mathrm{Sphere}}$ & $\mathcal{L}_{\mathrm{Sphere}}+\lambda_{\mathrm{SSAU}}\mathcal{L}_{\mathrm{SSAU}}$ & $\mathcal{L}_{\mathrm{Sphere}}$ \\
\midrule
4  & 8.03 & 9.42 & \textbf{5.38} & 5.02 & 24.71 & 25.67\\
6  & 7.22 & \textbf{9.10} & 5.36 & \textbf{5.04} & 24.73 & 25.80 \\		
8  & 7.01 & 9.41 & 5.33 & 5.00 & \textbf{24.74} & 25.79\\			
10  & 6.97 & 9.68 & 5.37 & 5.00 & 24.74 & 25.81\\		 			
12  & 6.97 & 10.01 & 5.32 & 5.00 & 24.74 & \textbf{25.83}\\					
14  & 6.93 & 10.07 & 5.33 & 4.97 & 24.72 & 25.83\\					
16  & \textbf{6.82} & 10.23 & 5.34 & 4.96 & 24.74 & 25.80\\		
\bottomrule
\end{tabular}
\end{table*}

\begin{table*}[t]
\centering
\caption{Ablation study of the proposed SSAU objective on the MS-COCO-2014 dataset.}
\label{contribution_align_loss_coco}
\begin{tabular}{c|cc|cc|cc}
\toprule
\multicolumn{7}{c}{\textit{Reconstruction Results (CFG/Step Independent)}} \\
\midrule
Metric & \multicolumn{3}{c|}{$\mathcal{L}_{\mathrm{Sphere}}+\lambda_{\mathrm{SSAU}}\mathcal{L}_{\mathrm{SSAU}}$ (Ours)} & \multicolumn{3}{c}{$\mathcal{L}_{\mathrm{Sphere}}$ (Original)} \\
\midrule
rFID $\downarrow$ & \multicolumn{3}{c|}{\textbf{0.78}} & \multicolumn{3}{c}{\textbf{0.78}} \\
rIS $\uparrow$    & \multicolumn{3}{c|}{\textbf{36.25}} & \multicolumn{3}{c}{36.24} \\
\midrule
\multicolumn{7}{c}{\textit{Generation Results (fixed steps=4)}} \\
\midrule
\multirow{2}{*}{CFG} &
\multicolumn{2}{c|}{gFID $\downarrow$} &
\multicolumn{2}{c|}{gIS $\uparrow$} &
\multicolumn{2}{c}{CLIPScore $\uparrow$} \\
& $\mathcal{L}_{\mathrm{Sphere}}+\lambda_{\mathrm{SSAU}}\mathcal{L}_{\mathrm{SSAU}}$ & $\mathcal{L}_{\mathrm{Sphere}}$ & $\mathcal{L}_{\mathrm{Sphere}}+\lambda_{\mathrm{SSAU}}\mathcal{L}_{\mathrm{SSAU}}$ & $\mathcal{L}_{\mathrm{Sphere}}$ & $\mathcal{L}_{\mathrm{Sphere}}+\lambda_{\mathrm{SSAU}}\mathcal{L}_{\mathrm{SSAU}}$ & $\mathcal{L}_{\mathrm{Sphere}}$ \\
\midrule
\textbf{2.0}  & \textbf{8.16} & \textbf{15.90} & 30.79 & \textbf{23.71} & 24.19 & \textbf{23.64}\\
3.0  & 9.00 & 27.53 & 31.87 & 20.63 & 24.52 & 23.03\\
4.0  & 9.49 & 41.08 & 32.43 & 17.79 & 24.75 & 22.24\\			
5.0  & 9.92 & 54.37 & 32.77 & 15.46 & 24.94 & 21.55\\					
6.0  & 10.24 & 65.41 & 33.01 & 13.87 & 25.11 & 21.00\\					
7.0  & 10.33 & 74.84 & \textbf{33.04} & 12.64 & 25.23 & 20.53\\					
8.0  & 10.38 & 82.26 & 32.93 & 11.73 & 25.36 & 20.13 \\					
9.0  & 10.54 & 88.09 & 32.87 & 10.98 & 25.45 & 19.82 \\					
10.0 & 10.67 & - & 32.88 & - & \textbf{25.54} & -\\
\midrule
\multicolumn{7}{c}{\textit{Generation Results (CFG scale=2.0)}} \\
\midrule
\multirow{2}{*}{CFG} &
\multicolumn{2}{c|}{gFID $\downarrow$} &
\multicolumn{2}{c|}{gIS $\uparrow$} &
\multicolumn{2}{c}{CLIPScore $\uparrow$} \\
& $\mathcal{L}_{\mathrm{Sphere}}+\lambda_{\mathrm{SSAU}}\mathcal{L}_{\mathrm{SSAU}}$ & $\mathcal{L}_{\mathrm{Sphere}}$ & $\mathcal{L}_{\mathrm{Sphere}}+\lambda_{\mathrm{SSAU}}\mathcal{L}_{\mathrm{SSAU}}$ & $\mathcal{L}_{\mathrm{Sphere}}$ & $\mathcal{L}_{\mathrm{Sphere}}+\lambda_{\mathrm{SSAU}}\mathcal{L}_{\mathrm{SSAU}}$ & $\mathcal{L}_{\mathrm{Sphere}}$ \\
\midrule
4  & 8.16 & \textbf{15.90} & \textbf{30.79} & \textbf{23.71} & \textbf{24.19} & \textbf{23.64} \\
\textbf{6} & \textbf{7.11} & 21.74 & 28.67 & 20.30 & 24.11 & 23.21\\
8 & 7.89 & 30.23 & 27.34 & 17.36 & 23.98 & 22.69\\
10  & 9.15 & 37.27 & 26.09 & 15.92 & 23.86 & 22.30\\	 			
12  & 10.38 & 43.81 & 25.42 & 14.65 & 23.73 & 21.92\\					
14  & 11.43 & 48.18 & 24.86 & 13.86 & 23.67 & 21.68\\					
16  & 12.34 & 51.95 & 24.56 & 13.39 & 23.58 & 21.48 \\		
\bottomrule
\end{tabular}
\end{table*}

\begin{table*}[t]
\centering
\caption{Ablation study of the proposed SSAU objective on the Flickr30k dataset.}
\label{contribution_align_loss_flickr}
\begin{tabular}{c|cc|cc|cc}
\toprule
\multicolumn{7}{c}{\textit{Reconstruction Results (CFG/Step Independent)}} \\
\midrule
Metric & \multicolumn{3}{c|}{$\mathcal{L}_{\mathrm{Sphere}}+\lambda_{\mathrm{SSAU}}\mathcal{L}_{\mathrm{SSAU}}$ (Ours)} & \multicolumn{3}{c}{$\mathcal{L}_{\mathrm{Sphere}}$ (Original)} \\
\midrule
rFID $\downarrow$ & \multicolumn{3}{c|}{0.82} & \multicolumn{3}{c}{\textbf{0.73}} \\
rIS $\uparrow$    & \multicolumn{3}{c|}{24.17} & \multicolumn{3}{c}{\textbf{24.38}} \\
\midrule
\multicolumn{7}{c}{\textit{Generation Results (fixed steps=4)}} \\
\midrule
\multirow{2}{*}{CFG} &
\multicolumn{2}{c|}{gFID $\downarrow$} &
\multicolumn{2}{c|}{gIS $\uparrow$} &
\multicolumn{2}{c}{CLIPScore $\uparrow$} \\
& $\mathcal{L}_{\mathrm{Sphere}}+\lambda_{\mathrm{SSAU}}\mathcal{L}_{\mathrm{SSAU}}$ & $\mathcal{L}_{\mathrm{Sphere}}$ & $\mathcal{L}_{\mathrm{Sphere}}+\lambda_{\mathrm{SSAU}}\mathcal{L}_{\mathrm{SSAU}}$ & $\mathcal{L}_{\mathrm{Sphere}}$ & $\mathcal{L}_{\mathrm{Sphere}}+\lambda_{\mathrm{SSAU}}\mathcal{L}_{\mathrm{SSAU}}$ & $\mathcal{L}_{\mathrm{Sphere}}$ \\
\midrule
2.0  & 17.79 & 17.84 & 16.74 & 17.38 & 23.01 & 23.25\\
3.0  & \textbf{17.22} & \textbf{17.74} & 18.35 & 18.20 & 23.43 & 23.76\\	
4.0  & 17.29 & 18.58 & 18.94 & \textbf{18.38} & 23.74 & 24.17\\	
5.0  & 17.38 & 18.93 & 19.49 & 18.37 & 23.98 & 24.49\\	
6.0  & 17.59 & 19.37 & 19.58 & 18.25 & 24.21 & 24.69\\					
7.0  & 17.53 & 19.63 & 19.63 & 17.89 & 24.37 & 24.85\\					
8.0  & 17.40 & 20.17 & 19.73 & 17.86 & 24.53 & 24.89\\					
9.0  & 17.58 & 20.61 & \textbf{19.82} & 17.20 & 24.65 & \textbf{24.91}\\					
10.0 & 17.59 & 21.01 & 19.46 & 17.04 & \textbf{24.73} & 24.87\\
\midrule
\multicolumn{7}{c}{\textit{Generation Results (CFG scale=3.0)}} \\
\midrule
\multirow{2}{*}{CFG} &
\multicolumn{2}{c|}{gFID $\downarrow$} &
\multicolumn{2}{c|}{gIS $\uparrow$} &
\multicolumn{2}{c}{CLIPScore $\uparrow$} \\
& $\mathcal{L}_{\mathrm{Sphere}}+\lambda_{\mathrm{SSAU}}\mathcal{L}_{\mathrm{SSAU}}$ & $\mathcal{L}_{\mathrm{Sphere}}$ & $\mathcal{L}_{\mathrm{Sphere}}+\lambda_{\mathrm{SSAU}}\mathcal{L}_{\mathrm{SSAU}}$ & $\mathcal{L}_{\mathrm{Sphere}}$ & $\mathcal{L}_{\mathrm{Sphere}}+\lambda_{\mathrm{SSAU}}\mathcal{L}_{\mathrm{SSAU}}$ & $\mathcal{L}_{\mathrm{Sphere}}$ \\
\midrule
4  & 17.22 & 17.74 & \textbf{18.35} & \textbf{18.20} & \textbf{23.43} & \textbf{23.76}\\
6  & 14.37 & \textbf{15.76} & 17.70 & 16.97 & 23.37 & 23.74\\	
8  & \textbf{13.86} & 16.13 & 17.20 & 16.36 & 23.27 & 23.71\\
10  & 14.26 & 17.15 & 16.78 & 15.55 & 23.27 & 23.63\\		 			
12  & 14.51 & 18.02 & 16.37 & 15.37 & 23.19 & 23.58\\					
14  & 14.90 & 18.54 & 16.10 & 15.09 & 23.14 & 23.51\\					
16  & 15.19 & 19.14 & 15.93 & 14.97 & 23.09 & 23.50\\		
\bottomrule
\end{tabular}
\end{table*}